# Fourier Feature Physics-Informed Neural Networks for Elasto-Plastic Analysis of Geomaterials with a Non-Associative Mohr-Coulomb Model

Apisit Robjanghvad and Sompote Youwai*

AI research group, Department of Civil Engineering, Faculty of Engineering, King Mongkut's University of Technology Thonburi, Thailand

*Corresponding author Email: sompote.you@kmutt.ac.th

## Abstract

Elasto-plastic boundary value problems in geotechnical engineering are conventionally solved by the Finite Element Method (FEM), which incurs high computational cost from incremental-iterative procedures. Physics-Informed Neural Networks (PINNs) offer a mesh-free alternative but suffer from spectral bias, preferentially fitting low-frequency components and failing to resolve the sharp gradients that arise at elastic-plastic boundaries and within localized plastic zones. This limitation is particularly consequential for the non-associative Mohr-Coulomb model, whose pressure-dependent yield surface and dilatant flow rule generate narrower plastic zones and steeper stress gradients than pressure-independent criteria. This study proposes a Fourier Feature Physics-Informed Neural Network (FF-PINN) for two-dimensional elasto-plastic problems governed by this model. Random Fourier feature mapping is embedded into the input layer to mitigate spectral bias, supported by a multi-objective loss function enforcing equilibrium, constitutive relations, and Karush-Kuhn-Tucker conditions against high-fidelity FEM data, together with a strain-adaptive sampling strategy. Benchmarked against the conventional PINN and FEM reference solutions across three test cases, FF-PINN achieves superior accuracy across most predicted fields, with error reductions of up to approximately 66 percent in displacement and approximately 27 percent in stress components, and reproduces the geometry of the plastic failure zones with markedly closer fidelity to FEM. Sensitivity analysis confirms robustness across training data size, collocation density, loss weighting, and noise levels up to 2.0 percent. FF-PINN converges in half the training epochs required by the conventional PINN, halving wall-clock training time while achieving greater predictive accuracy. The framework therefore offers a computationally efficient and physics-consistent alternative to FEM for elasto-plastic geotechnical analysis.

## 1. Introduction

The mechanical behavior of soils is fundamental to design and assessment of geotechnical structures, such as foundations, embankments, and slopes. To represent soil deformation and failure under loading, constitutive models based on elasto-plastic theory are extensively applied in geotechnical analyses (Chaboche, 2008). Among these models, the Mohr-Coulomb model is frequently adopted due to its simplicity and ability to represent the fundamental strength characteristics of soils (Yin et al., 2001; Zienkiewicz et al., 1975). Unlike metals, soils exhibit frictional and pressure-dependent behavior that causes associative flow rules to overestimate volumetric dilation; a non-associative flow rule is therefore required to accurately capture soil plastic deformation (Vermeer, 1998). The numerical solution of boundary value problems (BVPs) arising in geotechnical analysis has been predominantly addressed through the Finite Element Method (FEM) (Potts and Zdravković, 2001). FEM provides a robust and well-established framework for solving the governing equations of elasto-plastic deformation in complex geometries and boundary conditions. However, the inherently non-linear stress-strain behavior of soils requires an incremental-iterative solution procedure, such as the Newton-Raphson method, to be applied at each load increment, resulting in high computational cost and time consumption.

In recent years, Artificial Neural Networks (ANNs) have been increasingly applied in geotechnical engineering as data-driven surrogate models for predicting soil behavior and structural responses (Moayedi et al., 2020; W. Zhang et al., 2021). These approaches have addressed a wide range of problems, including the load-settlement modeling of axially loaded steel driven piles (Shahin, 2014), the bearing capacity of strip footings subjected to inclined loading (Kumar et al., 2024), the intelligent modeling of clay compressibility (P. Zhang et al., 2021), the stability analysis of rock and soil slopes (Millán and Galindo, 2025), and the load-deformation prediction of bored piles using transformer architectures (Youwai and Thongnoo, 2025). However, conventional ANNs operate as black-box models that require large, labeled datasets and cannot enforce the underlying physical laws, which limits their reliability and generalizability. To address these limitations, Physics-Informed Neural Networks (PINNs) were introduced as a framework that embeds governing equations, boundary conditions, and initial conditions directly into the loss function(Raissi et al., 2019). This mesh-free formulation bypasses iterative solvers and large datasets while solving the governing PDEs with greater flexibility than conventional numerical methods. Building on this framework, PINNs have been extended to a broad spectrum of solid mechanics problems spanning linear elasticity, hyperelasticity (Abueidda et al., 2022; Thakolkaran et al., 2022) and rate-independent plasticity under various yield criteria. A detailed review of related work is provided in Section 2.

Despite this progress, existing PINNs formulations for plasticity, including those addressing von Mises and Drucker Prager criteria, rely uniformly on standard fully connected architectures. Such architectures are known to exhibit spectral bias, whereby the network preferentially fits low frequency components of the solution and underresolves sharp local gradients (Rahaman et al., 2019). This limitation is particularly consequential for the non-associative Mohr Coulomb model considered here, since its pressure dependent yield surface and non-associative flow rule produces narrower, more sharply localized plastic zones and steeper stress gradients across the elastic plastic boundary than are typical of pressure independent criteria such as von Mises. Consequently, the very regions where accurate stress and strain resolution are most critical for practical geotechnical assessment, namely near yield boundaries and within localized plastic zones, are where a standard PINN is structurally least able to perform. Although Fourier feature mapping has been shown to mitigate spectral bias in other physics informed learning contexts (Tancik et al., 2020; Wang et al., 2021), it has not been examined for pressure dependent, non-associative plasticity, leaving open whether this remedy transfers to a constitutive setting governed by frictional strength and volumetric dilation rather than isotropic yielding. This gap motivates the present study.

To address this gap, this study proposes a Fourier Feature Physics-Informed Neural Network (FF-PINN) for solving elasto-plastic boundary value problems governed by the non-associative Mohr-Coulomb constitutive model. The main contributions are summarized as follows. First, the framework represents the first application of random Fourier feature mapping to pressure-dependent, non-associative plasticity within a PINNs setting. Second, the framework is shown to mitigate spectral bias and substantially improve the resolution of high-gradient displacement, stress, and plastic strain fields in localized plastic zones, yielding markedly higher stress component accuracy relative to the conventional PINN. Third, a multi-objective loss function is formulated to simultaneously enforce equilibrium equations, constitutive relations, and Karush-Kuhn-Tucker conditions against high-fidelity finite element reference data and is combined with a strain-adaptive sampling strategy that concentrates collocation points in regions of elevated strain to enhance accuracy in critical plastic transition zones. Fourth, the framework is rigorously benchmarked against conventional PINN and FEM reference solutions across three test cases with varying material parameters and loading conditions, and its robustness is established through a systematic sensitivity analysis covering training data size, collocation point density, loss weighting, and measurement noise; a computational cost analysis further demonstrates that FF-PINN achieves superior predictive accuracy across all fields while requiring substantially fewer training epochs than the conventional PINN.

The remainder of this paper is organized as follows. Section 2 reviews related PINNs studies in solid mechanics and plasticity and identifies the research gap addressed by the present work. Section 3 presents the mathematical formulation of the elasto-plastic problem, including the governing equations and the non-associative Mohr-Coulomb constitutive model. Section 4 describes the proposed FF-PINN framework, covering the conventional PINN baseline, the Fourier feature mapping, the multi-objective loss function, the strain-adaptive sampling strategy, and the error metrics. Section 5 reports three numerical benchmark examples, hyperparameter tuning, sensitivity analysis, and computational cost evaluation. Section 6 discusses the implications of the results, and Section 7 draws conclusions and outlines directions for future work.

## 2. Related work

PINNs, introduced by Raissi et al. (2019), have been progressively extended from linear elasticity (Chadha et al., 2023; Liang et al., 2023) to elasto plastic problems under von Mises (Chen et al., 2025; Haghighat et al., 2021; Roy and Guha, 2023) and Drucker Prager criteria (Roy et al., 2024), and further to elasto viscoplastic regimes (Eghtesad et al., 2024). Parallel developments have addressed related structural and geotechnical settings, including real time full field inference from sparse measurements (Go et al., 2025), nonlinear buried pipeline analysis under a Menegotto Pinto material model (Taraghi et al., 2025), and elasto-plastic analysis formulated in both strong and weak forms using multi network and Kolmogorov Arnold based architectures (Wang and Thai, 2026). Most recently, this line of work has been extended directly to Mohr Coulomb plasticity (Yuan et al., 2026; Zhang et al., 2026) and to pile soil interaction problems (Liu et al., 2026; Wu et al., 2026). Table 1 summarizes these studies with respect to yield criterion, network architecture, whether spectral bias mitigation is incorporated, sampling strategy, and reported limitations. As shown in Table 1, the sampling strategies adopted across these studies range from uniform and grid based sampling to random and adaptive schemes, yet this diversity in sampling has been pursued largely independently of the network architecture itself, with only Chadha et al. (2023) and Liang et al. (2023) for linear elasticity and Wang and Thai (2026) for von Mises plasticity incorporating any explicit mitigation of spectral bias (Rahaman et al., 2019). Neither of the two Mohr Coulomb specific formulations listed in Table 1 does so. Yuan et al. (2026) employs a multi network architecture with strain adaptive sampling yet reports lower precision for strain and difficulty resolving high gradient regions, while Zhang et al. (2026) report accuracy limited to homogeneous two-dimensional domains.

These reported limitations in Table 1 are consistent with the known behavior of spectrally biased networks, which preferentially learn low frequency components of a solution while underresolving sharp local gradients, and both Mohr Coulomb studies rely on standard architectures without any mechanism to address this behavior. This omission is significant for pressure dependent, non-associative plasticity in particular, since the frictional yield surface and non-associative flow rule of the Mohr Coulomb model generate narrower plastic zones and steeper stress gradients across the elastic plastic boundary than are typical of the pressure independent criteria for which spectral bias mitigation has so far been explored in Table 1. Fourier feature mapping is expected to address this limitation because it alters the spectral properties of the network at the level of the input representation rather than relying on training to overcome them. Neural tangent kernel analysis shows that a standard coordinate input induces a kernel whose eigenvalues decay rapidly with frequency, which is the root cause of spectral bias, whereas projecting the input coordinates through a bank of sinusoidal functions with a prescribed range of frequencies transforms this kernel into an approximately stationary form whose effective bandwidth can be tuned through the frequency scale of the mapping (Tancik et al., 2020). This gives the network access to high frequency basis functions from the outset of training rather than requiring it to acquire them gradually, and has been shown to improve convergence to functions with sharp local variation in other physics informed learning contexts (Wang et al., 2021). This property is expected to be particularly beneficial for the Mohr Coulomb problem, where the steep stress and strain gradients concentrated across the elastic plastic boundary correspond precisely to the high frequency content that a standard architecture is least able to represent. However, this expectation has not been examined for pressure dependent, non-associative plasticity, and its effectiveness cannot be assumed to transfer directly from isotropic to frictional, dilatant constitutive behavior, since the location and sharpness of the high gradient regions are themselves governed by the pressure sensitivity and non-associativity of the flow rule rather than fixed a priori. This gap, an unmitigated architectural limitation in the two most directly relevant prior studies on Mohr Coulomb plasticity, together with an unverified theoretical expectation of how Fourier feature mapping would perform in this setting, motivates the present study.

**Table 1** Summary of PINNs in Solid mechanics and Plasticity

| Reference | Yield Criterion | Architecture | Spectral Bias Mitigation | Sampling Strategy | Limitation |
|---|---|---|---|---|---|
| (Haghighat et al., 2021) | von-Mises | Multi-network DNNs | ✗ | Uniform / Random | Artificially diffuse sharp gradients |
| (Chadha et al., 2023) | Linear Elastic | DEM (MLP) with Two-loop optimization | ✓ | Uniform (Grid-based) | High hyperparameter sensitivity; load-dependent accuracy |
| (Liang et al., 2023) | Linear Elastic | PINNFD (FCNN) | ✓ | Random | Local errors near loading point |
| (Roy and Guha, 2023) | von-Mises | Multiple Dense ANNs | ✗ | Uniform | Computationally slower than FEM; errors at stress concentrations |
| (Roy et al., 2024) | Drucker-Prager | Independent Dense FFNNs | ✗ | Uniform (Grid-based) | Long training time; localized errors near sharp features |
| (Chen et al., 2025) | von-Mises | PiNet (CPM vs. DRM) | ✗ | Uniform / Adaptive | DRM poor for stress/strain; sensitive to noise |
| (Go et al., 2025) | None (Linear elastic) | SA-PINN | ✗ | Uniform | Extremely long training time; noise sensitivity |
| (Taraghi et al., 2025) | Menegotto–Pinto | FCNN | ✗ | Random (LHS) | Restricted to global deformation; cannot capture local buckling |
| (Liu et al., 2026) | None | MLP-PINN + FEM | ✗ | Random | Error accumulation; overfitting risk |
| (Wang and Thai, 2026) | von-Mises | Independent multi-networks (MLP/KAN) | ✓ | Random | Long training time; residual imbalance |
| (Wu et al., 2026) | None (Linear elastic) | TLP-mPINNs | ✗ | Weighted / Random | 1D pile simplification; hyperparameter sensitivity |
| (Yuan et al., 2026) | Mohr–Coulomb (M–C) (Assoc. / Non-assoc.) | Multi-DNN (6 nets) | ✗ | Uniform / Strain Adaptive (SAS) | Lower precision for strain; struggles with high gradients |
| (Zhang et al., 2026) | Mohr–Coulomb (M–C) (Associated Flow Rule) | Adaptive PINN (MLP) | ✗ | Global Adaptive | Slower than FEM; limited to homogeneous 2D |

## 3. Mathematical Modeling for Geotechnical Materials

### 3.1. Governing Equations

In this study, the deformation of the soil mass is assumed to be small. Under the small deformation assumption, the equilibrium equation can be written as:

$$\sigma_{ij,j} + f_i = 0 \tag{1}$$

where $\sigma_{ij}$ denotes the Cauchy stress tensor, describing the internal normal and shear stresses acting at a point within the medium, while $f_i$ represents the body force. Under the assumption of small deformations, the strain-displacement relation is defined as follows:

$$\varepsilon_{ij} = \frac{1}{2}\left(u_{i,j} + u_{j,i}\right) \tag{2}$$

where $\varepsilon_{ij}$ denotes the strain tensor and $u_i$ represents the displacement field.

For linear elasticity, the Cauchy stress tensor $\sigma$ can be calculated by the constitutive relation as follows:

$$\sigma_{ij} = \lambda\delta_{ij}\varepsilon_{kk} + 2\mu\varepsilon_{ij} \tag{3}$$

where $\lambda$ and $\mu$ represent the Lamé constants, and $\delta_{ij}$ represents the Kronecker delta. For an elasto-plastic material, the total strain tensor can be expressed as the sum of elastic $\varepsilon_{ij}^{e}$ and plastic strains $\varepsilon_{ij}^{p}$.

$$\varepsilon_{ij} = \varepsilon_{ij}^{e} + \varepsilon_{ij}^{p} \tag{4}$$

### 3.2. Non-associative Mohr-Coulomb Constitutive Model

To implement the Mohr-Coulomb model, we adopt the Menétrey–Willam yield function (Menétrey and Willam, 1995) to describe the pressure-dependent failure surface. Fig. 1 shows the adopted Menétrey-Willam representation of the Mohr-Coulomb yield surface in two complementary sections. Panel (a) is the meridional section, plotting modified deviatoric stress against mean stress. The yield condition traces a straight line with the cohesion as intercept and the friction angle governing the slope, closing at the apex that marks the tensile limit. States below the line are elastic and states on it are yielding. This inclined trace embodies the pressure dependence absent from pressure-independent criteria. Panel (b) is the deviatoric section, viewed along the hydrostatic axis with the principal stress axes projected radially and position defined by the deviatoric polar angle and radius. Four traces are superimposed: the circular von Mises surface as outer bound, the irregular hexagonal classical Mohr-Coulomb surface whose singular corners impede numerical implementation, the smooth elliptic Menétrey-Willam surface adopted here, which coincides with the hexagon at the compression corners while remaining continuously differentiable between them, and the triangular Rankine limiting case. The eccentricity parameter controls the rounding, recovering the circular form in the limit. In this study, the constitutive model is developed in total form rather than the commonly used rate form. The yield surface is then written as:

$$F = R_{mc}q - p\tan\phi - c = 0 \tag{5}$$

where $p$ is the mean stress, $q$ is the deviatoric stress, $\phi$ is the friction angle, and $c$ is the cohesion. The term $R_{mc}(\Theta, \phi)$ is the Mohr-Coulomb elliptic function used to describe the shape of the yield surface in the deviatoric plane, expressed as:

$$R_{mc}(\Theta,\phi) = \frac{1}{\sqrt{3}\cos\phi}\sin\left(\Theta+\frac{\pi}{3}\right) + \frac{1}{3}\cos\left(\Theta+\frac{\pi}{3}\right)\tan\phi \tag{6}$$

where $\Theta$ is the deviatoric polar angle (Chen and Han, 2007) defined as

$$\cos(3\Theta) = \left(\frac{r}{q}\right)^3 \tag{7}$$

The stress invariants $p$, $q$, and $r$ are defined in terms of the stress tensor component as follows:

$$p = -1/3(\sigma_{11} + \sigma_{22} + \sigma_{33}) \tag{8}$$

$$q = \sqrt{\frac{3}{2}S_{ij}S_{ij}} \tag{9}$$

$$r = \left(\frac{9}{2}S_{ij}S_{jk}S_{ki}\right)^{1/3} \tag{10}$$

where $S_{ij}$ is the deviatoric stress tensor, given by $S_{ij} = \sigma_{ij} + p\,\delta_{ij}$.

To accurately model the dilatancy behavior of geomaterials, a non-associative flow rule is employed with the plastic potential function G defined independently from the yield function F. Following (Menétrey and Willam, 1995), the plastic potential is expressed as:

$$G = \sqrt{(\epsilon c|_0 \tan\psi)^2 + (R_{mw}q)^2} - p\tan\psi \tag{11}$$

where $\psi$ is the dilatancy angle, $c|_0$ is the initial cohesion yield stress, and $\epsilon$ is a meridional eccentricity that determines the rate at which the flow potential approaches its asymptote. The elliptic function $R_{mw}(\Theta, e)$ ensures smoothness in the deviatoric plane:

$$R_{mw}(\Theta,e) = \frac{4(1-e^2)\cos^2\Theta + (2e-1)^2}{2(1-e^2)\cos\Theta + (2e-1)\sqrt{4(1-e^2)\cos^2\Theta + 5e^2 - 4e}} R_{mc}\left(\frac{\pi}{3},\phi\right) \tag{12}$$

where $e$ is the eccentricity parameter and $R_{mc}\left(\frac{\pi}{3},\phi\right) = \frac{3-sin(\phi)}{6cos(\phi)}$. For the PINN implementation, the total form of the non-associated flow rule is employed to define the plastic strain constraint. The plastic strain tensor $\epsilon^p$ is expressed as:

$$\varepsilon^p = \lambda\frac{\partial G}{\partial\sigma} \tag{13}$$

where $\lambda$ is the plastic multiplier, which is explicitly calculated to satisfy the Karush-Kuhn-Tucker (KKT) optimality conditions and the consistency requirement, ensuring that the stress state remains on or within the yield surface.

The Karush-Kuhn-Tucker (KKT) conditions, which govern the activation of plastic flow (Simo and Hughes, 1998), are expressed as follows:

$$\lambda \geq 0, \quad F(\sigma) \leq 0, \quad \lambda F(\sigma) = 0 \tag{14}$$

where $F(\boldsymbol{\sigma})$ is the Mohr-Coulomb yield function. These conditions ensure that plastic strain occurs only when the stress state lies on the yield surface ($F = 0$), and remains zero during elastic loading ($F < 0$).

To complete the elasto-plastic constitutive framework, the total stress tensor σ is related to the total strain tensor ε through the elastic stiffness matrix $C^e$, accounting for the plastic strain defined in Eq. (13):

$$\sigma = C^e : (\varepsilon - \varepsilon^p) \tag{15}$$

where $C^e$ is the fourth-order isotropic elastic stiffness tensor defined by the Lamé constants ($\lambda_{\text{Lamé}}$) and $\mu$ as introduced in Eq. (3). The plastic multiplier $\lambda$ is determined explicitly by substituting Eq. (15) into the consistency condition $F(\sigma) = 0$ and solving directly, yielding:

$$\lambda = \frac{\frac{\partial F}{\partial \sigma} : C^e : \varepsilon}{\frac{\partial F}{\partial \sigma} : C^e : \frac{\partial G}{\partial \sigma}} \tag{16}$$

where $\frac{\partial F}{\partial \sigma}$ and $\frac{\partial G}{\partial \sigma}$ denote the gradient of the yield function and plastic potential function with respect to the stress tensor, respectively. The value of $\lambda$ is subject to the KKT conditions stated in Eq. (14), ensuring that plastic flow is activated only when the stress state lies on the yield surface ($F = 0$), while $\lambda = 0$ during the elastic loading ($F < 0$)

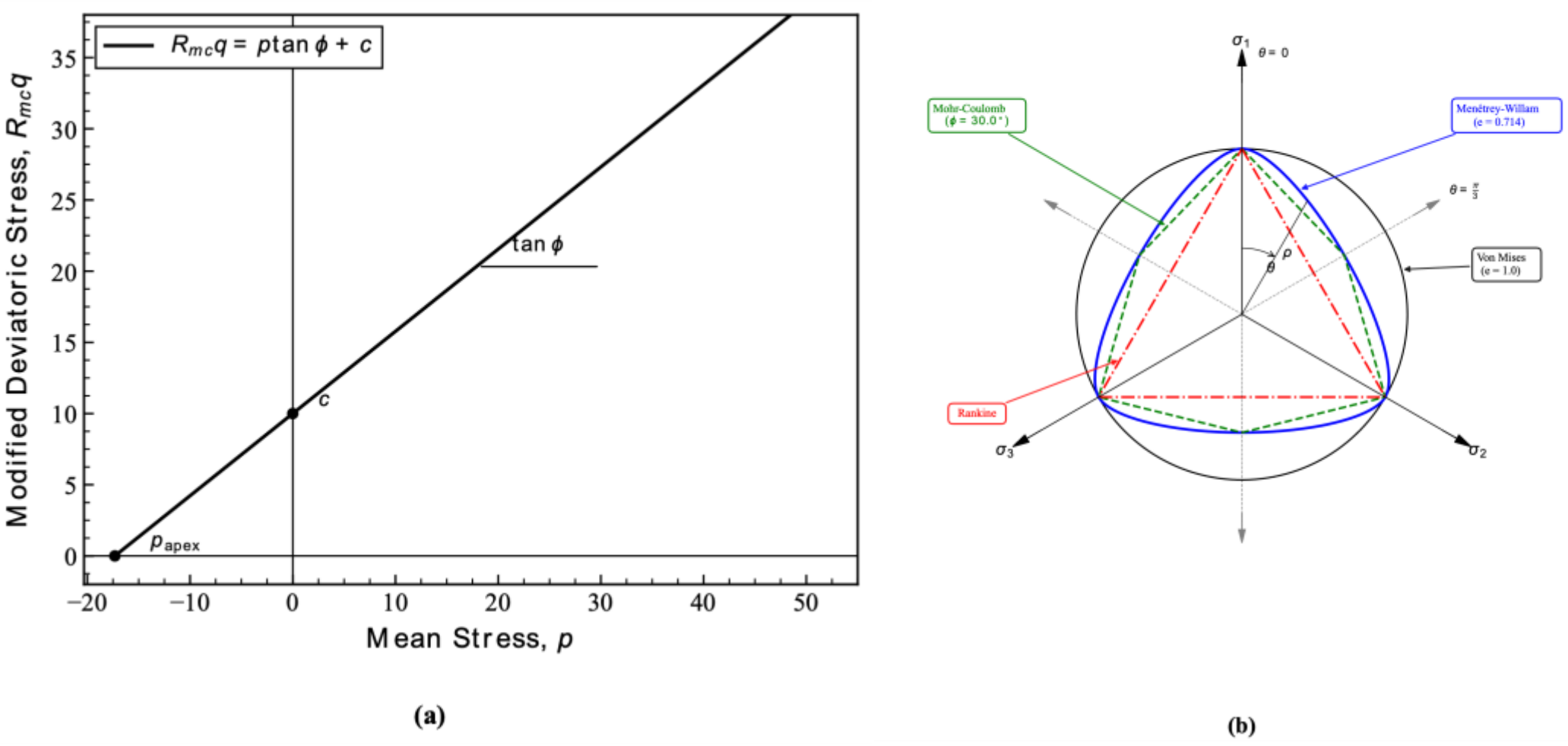

**Fig. 1.** Schematic of (a) Mohr-Coulomb yield surface in meridional planes; and (b) flow potential in the deviatoric plane. (Redrawn from Dassault Systèmes Simulia, 2016)

## 4. PINNs architectures setup

### 4.1 Conventional Physics-Informed Neural Networks (PINNs)

Physics-Informed Neural Networks (PINNs), a class of deep learning frameworks, integrate governing equations such as partial differential equations (PDEs) and boundary conditions directly into the loss function. This framework was introduced by Raissi et al. (2019) to solve both forward and inverse problems involving nonlinear partial differential equations. In the field of solid mechanics, this approach has been specifically extended to solve elasto-plastic problems. For instance, Haghighat et al. (2021) and Roy et al. (2024) developed PINNs frameworks to model von Mises and Drucker-Prager plasticity, respectively. In these applications, a feed-forward neural network $N(x;\theta)$ is employed to simultaneously approximate the displacement and stress fields, where $x = (x, y)$ denotes the spatial coordinate and $\theta$ represents the set of trainable parameters. The network maps the input coordinate to the output variable as:

$$\mathcal{N}: [x; y] \mapsto \left[u(x,y); v(x,y); \sigma_{xx}(x,y); \sigma_{yy}(x,y); \sigma_{zz}(x,y); \sigma_{xy}(x,y)\right] \quad (17)$$

where $u$ and $v$ are the displacement components in the $x$ and $y$ direction, respectively, and $\sigma_{xx}$, $\sigma_{yy}$, $\sigma_{zz}$, and $\sigma_{xy}$ are the in-plane stress components. The networks consist of an input layer, $L$-1 hidden layers, and an output layer. The forward pass through each layer is expressed as:

$$\begin{gathered} z^{(0)} = x = [x; y] \\ z^{(k)} = \sigma\left(W^{(k)} z^{(k-1)} + b^{(k)}\right), \quad k = 1,2,\dots,L-1 \\ \hat{u} = z^{(L)} = W^{(L)} z^{(L-1)} + b^{(L)} \end{gathered} \quad (18)$$

where $W^{(k)} \in \mathbb{R}^{n_k \times n_{k-1}}$ and $b^{(k)} \in \mathbb{R}^{n_k}$ are the weight matrix and bias vector of the $k$-th layer, respectively, $n_k$ denotes the number of neurons in layer $k$, and $\sigma(\cdot)$ is the activation function. In PINN-based solid mechanics applications, smooth and infinitely differentiable activation functions such as tanh (Glorot and Bengio, 2010) and swish are preferred, as the governing equation requires higher-order spatial derivatives of the network's output. The full set of trainable parameters is defined as $\theta = \left\{W^{(k)}, b^{(k)}\right\}_{k=1}^{L}$, which are optimized by minimizing a composite loss function encoding the residuals of the governing PDEs, boundary conditions, and available data, as described in the following section.

Fig. 2 summarizes the conventional PINN architecture and its optimization workflow. The spatial coordinates enter a fully connected network of hidden layers with tanh activation, which outputs the two displacement components and the four in-plane stress components. Automatic differentiation of these outputs with respect to the coordinates supplies the displacement and stress gradients required to form the strain field and the equilibrium residual. These quantities assemble into the total loss, comprising the data misfit against the finite element reference across displacement, stress, total strain, and plastic strain, together with the weighted physics loss combining the equilibrium, constitutive, Karush-Kuhn-Tucker, and boundary residuals. The total loss is tested against a convergence tolerance: while it exceeds the threshold, the network parameters are updated by minimization and the cycle repeats; once it falls below, training terminates.

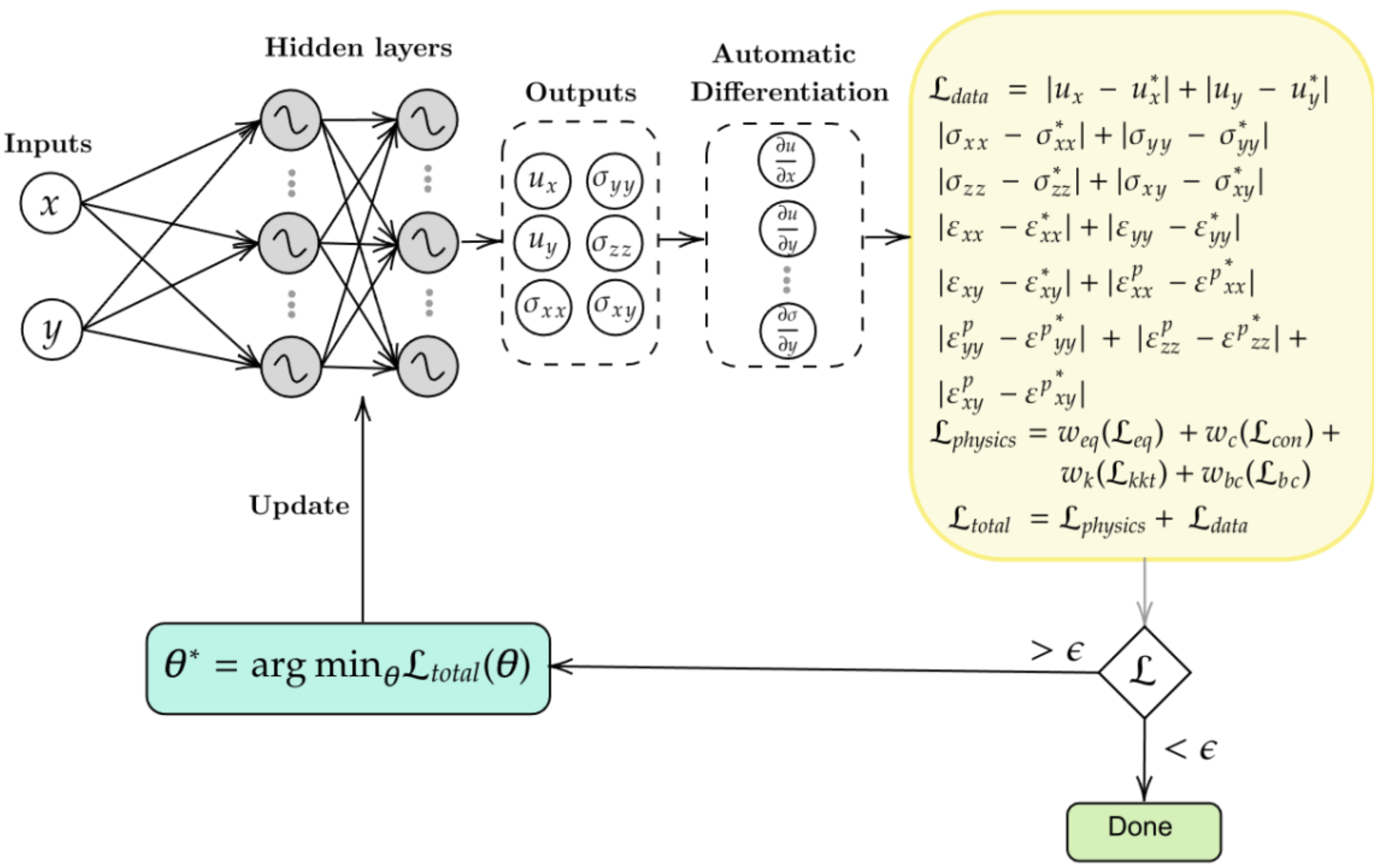

**Fig. 2**. Schematic diagram of the PINNs architecture and the optimization workflow for elasto-plastic analysis.

### 4.2 Fourier Feature Mapping

Although a conventional FCNN can approximate continuous functions, it exhibits spectral bias, preferentially learning low-frequency components while failing to capture high-frequency variations (Rahaman et al., 2019; Tancik et al., 2020; Xu et al., 2019, 2025). To address this limitation, a Fourier feature mapping applies sine and cosine transformations to the input coordinate $(x, y)$ before they are passed into the hidden layers, defined as:

$$\gamma(v) = [\cos(2\pi Bv), \sin(2\pi Bv)]^T \tag{19}$$

where $\boldsymbol{v} = (x, y) \in \mathbb{R}^2$ denotes the input spatial coordinates, $B \in \mathbb{R}^{(m \times d)}$ contains random frequencies, $m$ is the number of Fourier features, and $d = 2$ is the input dimension. The entries of $\boldsymbol{B}$ are sampled independently from a Gaussian distribution:

$$B_{ij} \sim \mathcal{N}(0, \sigma^2), i = 1, \ldots, m, j = 1, \ldots, \tag{20}$$

where σ is a scale hyperparameter controlling the frequency bandwidth. The matrix $\boldsymbol{B}$ is sampled once prior to training and remains fixed throughout optimization; it is therefore excluded from the trainable parameter set θ.

The mapped featured vector $\gamma(v) \in \mathbb{R}^{2m}$ replaces the raw coordinate input in Eq. (18), serving as the first-layer input to the subsequent hidden layers:

$$z^{(0)} = \gamma(v) \in \mathbb{R}^{2m} \tag{21}$$

The remaining forward-pass computations follow the same formulation as defined in Eq. (18). The result of the FF-PINN architecture is illustrated in Fig. 3.

Fig. 3 shows the proposed FF-PINN architecture, which differs from Fig. 2 solely in the treatment of the input layer. Rather than passing the spatial coordinates directly to the hidden layers, the coordinates are first projected through a random Fourier feature mapping, which lifts them into a higher-dimensional set of sinusoidal basis functions with frequencies drawn from a fixed Gaussian distribution. The mapped vector then serves as the first-layer input to the same fully connected network. The frequency matrix is sampled once before training and held fixed, so it adds no trainable parameters. All downstream components remain unchanged: the network outputs the displacement and in-plane stress components, automatic differentiation supplies the required gradients, and the same composite loss and convergence-tested update loop govern training. Isolating the modification to the input representation ensures that any difference in performance relative to the conventional PINN is attributable to the spectral properties of the encoding rather than to network capacity or optimization settings.

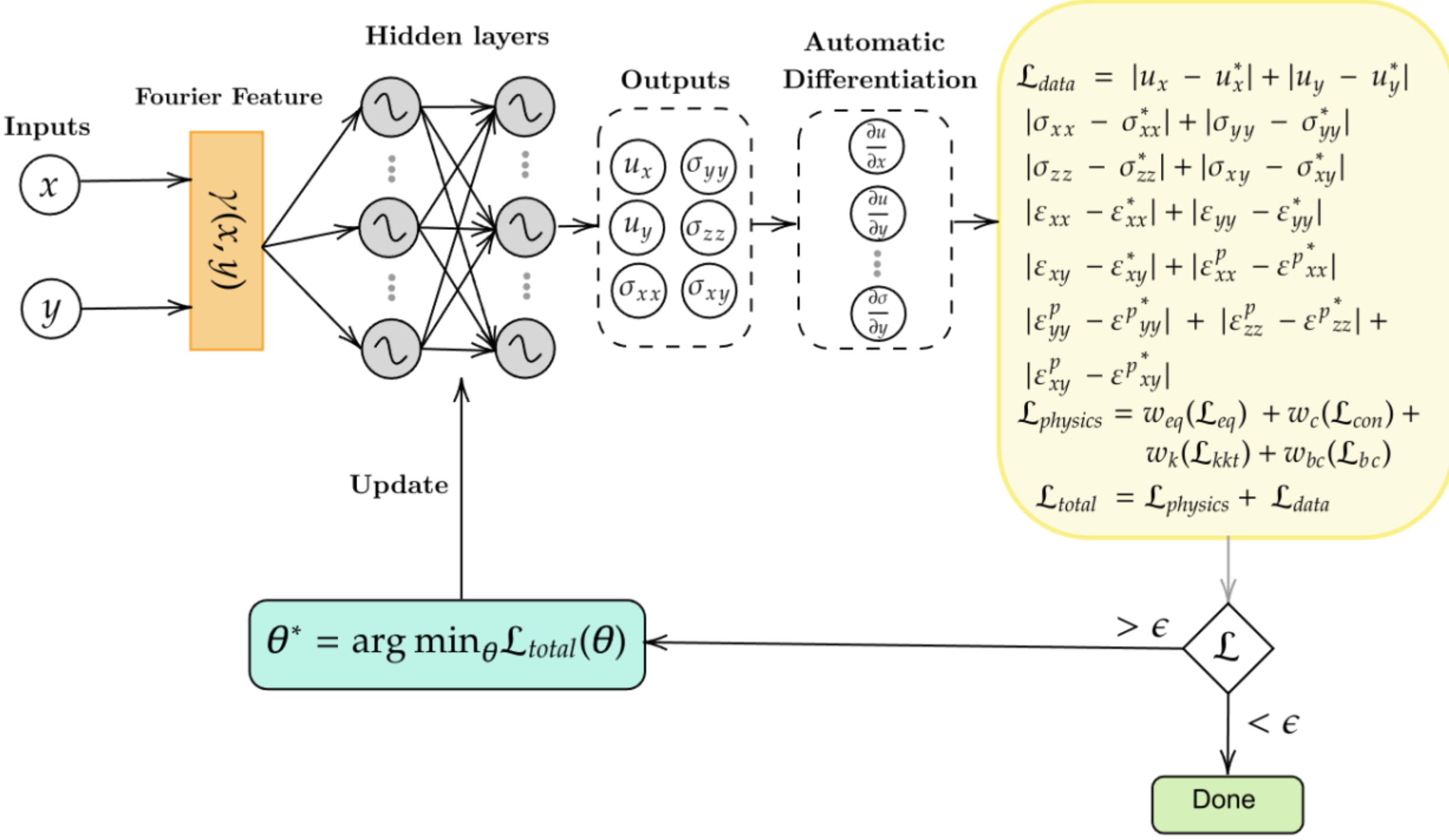

**Fig. 3**. Schematic diagram of the FF-PINN architecture incorporating a Fourier feature mapping to enhance the capture of complex nonlinearities.

### 4.3 Multi-objective loss function

The multi-objective loss function presented throughout this section is applicable to both the conventional PINN (Section 4.1) and the FF-PINN (Section 4.2) architectures, with all loss formulations defined with respect to the network output $\hat{u} = z^{(L)}$ as described in Eqs. (17)-(18) and (21). In contrast to conventional purely data-driven neural networks, which operate as black-box models and predict responses without inherent physical constraints, PINN integrates governing equations directly into the learning process. By embedding the underlying physical laws, such as equilibrium equations and constitutive relations, into the loss function, PINN ensures that the model remains physically consistent, even in regimes where training data is sparse or noisy. In elasto-plastic analysis, the network must respect the equilibrium equation, the Mohr-Coulomb yield criterion, the Karush-Kuhn-Tucker (KKT) conditions, and various boundary conditions.

### 4.3.1 Physics loss

The physics-based loss function $\mathcal{L}_{physics}$ enforces the underlying governing equation and boundary conditions at a set of collocation points sampled within the problem domain. It is composed of four weighted residual terms:

$$\mathcal{L}_{physics} = w_{eq}\mathcal{L}_{eq} + w_c\mathcal{L}_{con} + w_{KKT}\mathcal{L}_{KKT} + w_{bc}\mathcal{L}_{bc} \tag{22}$$

Following the equilibrium equation in Eq. (1), the equilibrium residual can be expressed as

$$\mathcal{L}_{eq} = \frac{1}{N}\sum_{i=1}^{N} ||\sigma_{ij,j}(x_i) + f_i(x_i)||^2 \tag{23}$$

where $f_i = 0$ denotes the absence of body forces and $N$ represents the total number of collocation points distributed within the domain for evaluating the physics residual. The constitutive loss $\mathcal{L}_{con}$ is introduced to enforce the stress-strain relation defined in Eq. (15), which can be formulated as:

$$\mathcal{L}_{con} = \frac{1}{N}\sum_{i=1}^{N}||\sigma_i^{NN} - C^e: (\varepsilon - \varepsilon^p)||^2 \tag{24}$$

where $\sigma_i^{NN}$ is the stress tensor predicted by the neural network at the collocation point $i$, $C^e$ is the elastic stiffness tensor, $\varepsilon$ is the total strain tensor computed from the displacement gradient via Eq. (2), and $\varepsilon^p$ is the plastic strain defined in Eq. (13). While the constitutive loss $\mathcal{L}_{con}$ enforces the stress-strain relationship, it does not explicitly constrain the activation of plastic flow. To ensure that the predicted stress satisfies the elasto-plastic switching conditions governed by the KKT inequalities in Eq. (14), the KKT loss $\mathcal{L}_{KKT}$ is introduced as:

$$\mathcal{L}_{KKT} = \frac{1}{N}\sum_{i=1}^{N}[\langle -\lambda_i\rangle^2 + \langle F_i\rangle^2 + (\lambda_i F_i)^2] \tag{25}$$

where $\langle\cdot\rangle = \max(\cdot, 0)$ denotes the Macaulay bracket. The first term penalizes any negative values of the plastic multiplier, enforcing the $\lambda \geq 0$. The second term penalizes stress states that violate the yield surface, enforcing $F(\sigma) \leq 0$. The third term enforces complementary conditions $\lambda F(\sigma) = 0$, ensuring that plastic flow is activated only when the stress state lies on the yield surface (F = 0), while $\lambda$ remains zero during elastic loading (F < 0). The boundary loss $\mathcal{L}_{bc}$ is decomposed into two terms corresponding to the Dirichlet and Neumann boundary conditions:

$$\mathcal{L}_{bc} = \mathcal{L}_{bc}^{D} + \mathcal{L}_{bc}^{N} \tag{26}$$

The Dirichlet loss term, $\mathcal{L}_{bc}^{D}$, ensures that the neural network predictions satisfy the prescribed displacement conditions on the essential boundaries, expressed as:

$$\mathcal{L}_{bc}^{D} = \frac{1}{N_b}\sum_{i=1}^{N_b} ||u_i^{NN} - \overline{u_i}||^2 \tag{27}$$

Similarly, the Neumann loss term, $\mathcal{L}_{bc}^{N}$, enforces the prescribed traction boundary conditions:

$$\mathcal{L}_{bc}^{N} = \frac{1}{N_b}\sum_{i=1}^{N_b} ||\sigma_i^{NN} \cdot n_i - \overline{t_i}||^2 \tag{28}$$

In these formulations, $N_b$ denotes the number of collocation points along the boundaries. The variables $u_i^{NN}$ and $\overline{u_i}$ represent the predicted and prescribed displacement vectors, respectively. Furthermore, $\sigma_i^{NN}$, $n_i$, and $\overline{t_i}$ correspond to the predicted stress tensor, the outward unit normal vector, and the prescribed traction vector at each boundary point $i$.

### 4.3.2 Data-driven loss

To further guide the network training with reference solutions obtained from high-fidelity FEM solutions or experimental data, the data-driven loss $\mathcal{L}_{data}$ is introduced to penalize the discrepancy between the network and the reference data, across displacement, stress, strain fields, and plastic strain:

$$\mathcal{L}_{data} = \mathcal{L}_u + \mathcal{L}_\sigma + \mathcal{L}_\varepsilon + \mathcal{L}_{\varepsilon^p} \tag{29}$$

$$\mathcal{L}_u = \frac{1}{N_d}\sum_{i=1}^{N_d} ||u_i^{NN} - u_i^{ref}||^2 \tag{30}$$

$$\mathcal{L}_\sigma = \frac{1}{N_d}\sum_{i=1}^{N_d} ||\sigma_i^{NN} - \sigma_i^{ref}||^2 \tag{31}$$

$$\mathcal{L}_\varepsilon = \frac{1}{N_d}\sum_{i=1}^{N_d} ||\varepsilon_i^{NN} - \varepsilon_i^{ref}||^2 \tag{32}$$

$$\mathcal{L}_{\varepsilon^p} = \frac{1}{N_d}\sum_{i=1}^{N_d} ||{\varepsilon^p}_i^{NN} - {\varepsilon^p}_i^{ref}||^2 \tag{33}$$

where $u_i^{NN}$ and $\sigma_i^{NN}$ denote the network-predicted displacement and stress tensor at data point $i$, $\varepsilon_i^{NN} = \nabla {u_i}^{NN}$ denotes the predicted strain tensor derived from the network-predicted displacement via automatic differentiation, and $u_i^{ref}$, $\sigma_i^{ref}$, and $\varepsilon_i^{ref}$ denote the corresponding reference solutions obtained from high-fidelity FEM, evaluated over $N_d$ data points. In contrast to the preceding fields, the predicted plastic strain ${\varepsilon^p}_i^{NN}$ is not a direct network output but is derived from the network-predicted stress $\sigma_i^{NN}$ through the non-associated flow rule and plastic multiplier defined in Eqs. (13) and (16), respectively. The loss $\mathcal{L}_{\varepsilon^p}$ is evaluated only at data points where $|{\varepsilon^p}_i^{ref}| > \delta$, to avoid bias from the dominant elastic region where ${\varepsilon^p}_i^{ref} = 0$.

### 4.3.3 Combined Physics and Data-Driven Loss

We formulate the total loss function as a weighted combination of the physics-based loss defined in Section 4.3.1 and the data-driven loss introduced in Section 4.3.2, enabling the network to simultaneously satisfy the governing physical laws and conform to reference solutions obtained from high-fidelity FEM simulations:

$$\mathcal{L}_{total} = w_{physics}\mathcal{L}_{physics} + w_{data}\mathcal{L}_{data}$$

$$\mathcal{L}_{total} = w_{physics}\left(w_{eq}\mathcal{L}_{eq} + w_{c}\mathcal{L}_{con} + w_{KKT}\mathcal{L}_{KKT} + w_{bc}\mathcal{L}_{bc}\right) \quad (34)$$

$$+w_{data}(\mathcal{L}_{u} + \mathcal{L}_{\sigma} + \mathcal{L}_{\varepsilon} + \mathcal{L}_{\varepsilon^{p}})$$

where $w_{physics}$ and $w_{data}$ are the global weighting coefficients governing the relative contribution of each loss component during training. The individual sub-weights $w_{eq}$, $w_{c}$, $w_{KKT}$, and $w_{bc}$ are defined as introduced in Eq. (22). In this study, the global loss weights are set to $w_{physics}$ = 1 and $w_{data}$ = 15. The relatively high weight assigned to the data-driven loss reflects the importance of anchoring the network predictions to high-fidelity FEM reference solutions, particularly in the localized plastic zone where the physics residuals alone are insufficient to achieve convergence.

## 4.4 Sampling technique

In this study, we employ two sampling techniques to generate training data points within the domain, as illustrated in Fig. 4. Grid sampling provides a uniform distribution of points across the domain, whereas Strain Adaptive Sampling (Yuan et al., 2026) yields a higher density of data points in regions of elevated strain magnitudes. The strain components $\varepsilon_{xx}^{i}$, $\varepsilon_{yy}^{i}$ and $\varepsilon_{xy}^{i}$ are extracted from high-fidelity FEM simulations at each sampling point $i$, and the local strain magnitude $S_i$ is defined as:

$$S_i = \sqrt{\left(\varepsilon_{xx}^{i}\right)^2 + \left(\varepsilon_{yy}^{i}\right)^2 + \left(\varepsilon_{xy}^{i}\right)^2} \quad (35)$$

The selection probability of each candidate point is then proportional to the strain magnitude, given by

$$P_i = \frac{S_i}{\sum_i S_i} \quad (36)$$

where the summation is taken over all candidate points in the domain, ensuring that regions with higher strain are sampled more frequently.

Fig. 4 contrasts the spatial distribution of training data points produced by the two strategies. In panel (a), grid sampling places points on a regular lattice, so density is independent of the solution field and the sharp elastic-plastic transition beneath the loaded strip receives no more representation than the smooth far field. In panel (b), strain-adaptive sampling yields a strongly heterogeneous distribution, with points clustering beneath and adjacent to the loaded strip where the plastic zone develops, and thinning toward the lower-right of the domain, which remains elastic. At fixed total point count, this reallocates training effort from regions the network already resolves easily to those where the solution varies most rapidly.

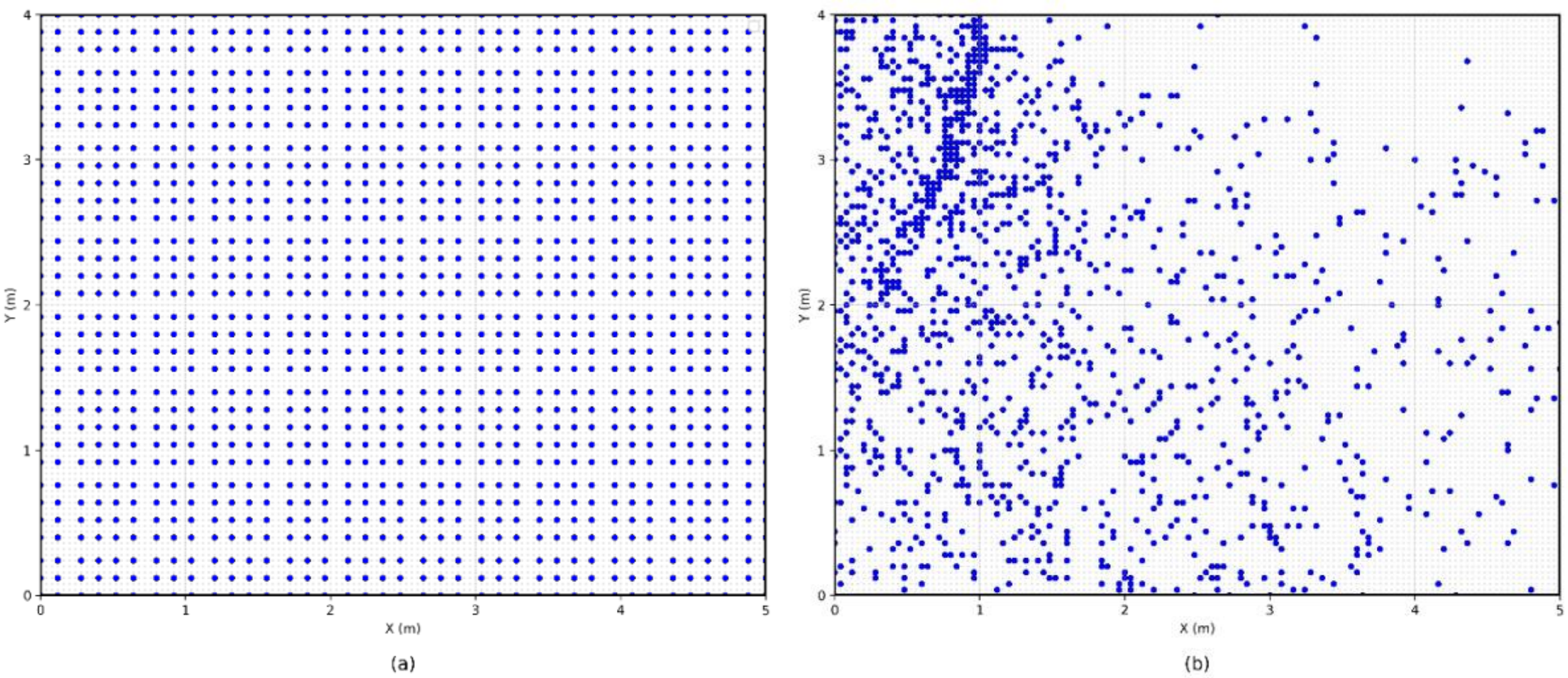

**Fig. 4.** Spatial distribution of training data points: (a) structured grid sampling and (b) strain-adaptive sampling.

### 4.5 Error Analysis

To assess the accuracy of the model, the point-wise relative error is calculated for each field variable (displacement, stress, and strain) for the predictions, which is defined as:

$$\epsilon^{rel} = \frac{\left|y_i^{ref} - y_i^{NN}\right|}{\left|max(y_i^{ref})\right|} \tag{37}$$

where $y_i^{ref}$ denotes the reference FEM solution and $y_i^{NN}$ denotes the prediction from the PINN model at point $i$. Furthermore, to evaluate the overall performance and global accuracy of the model across the entire computational domain, formulated as:

$$\text{Relative } L_2 \text{ Error} = \frac{\sqrt{\sum_{i=1}^{N}\left(y_i^{NN} - y_i^{ref}\right)^2}}{\sqrt{\sum_{i=1}^{N}\left(y_i^{ref}\right)^2}} \tag{38}$$

Where $N$ is the total number of evaluation data points distributed within the domain.

## 5. Numerical Examples

### 5.1 Problem Definition

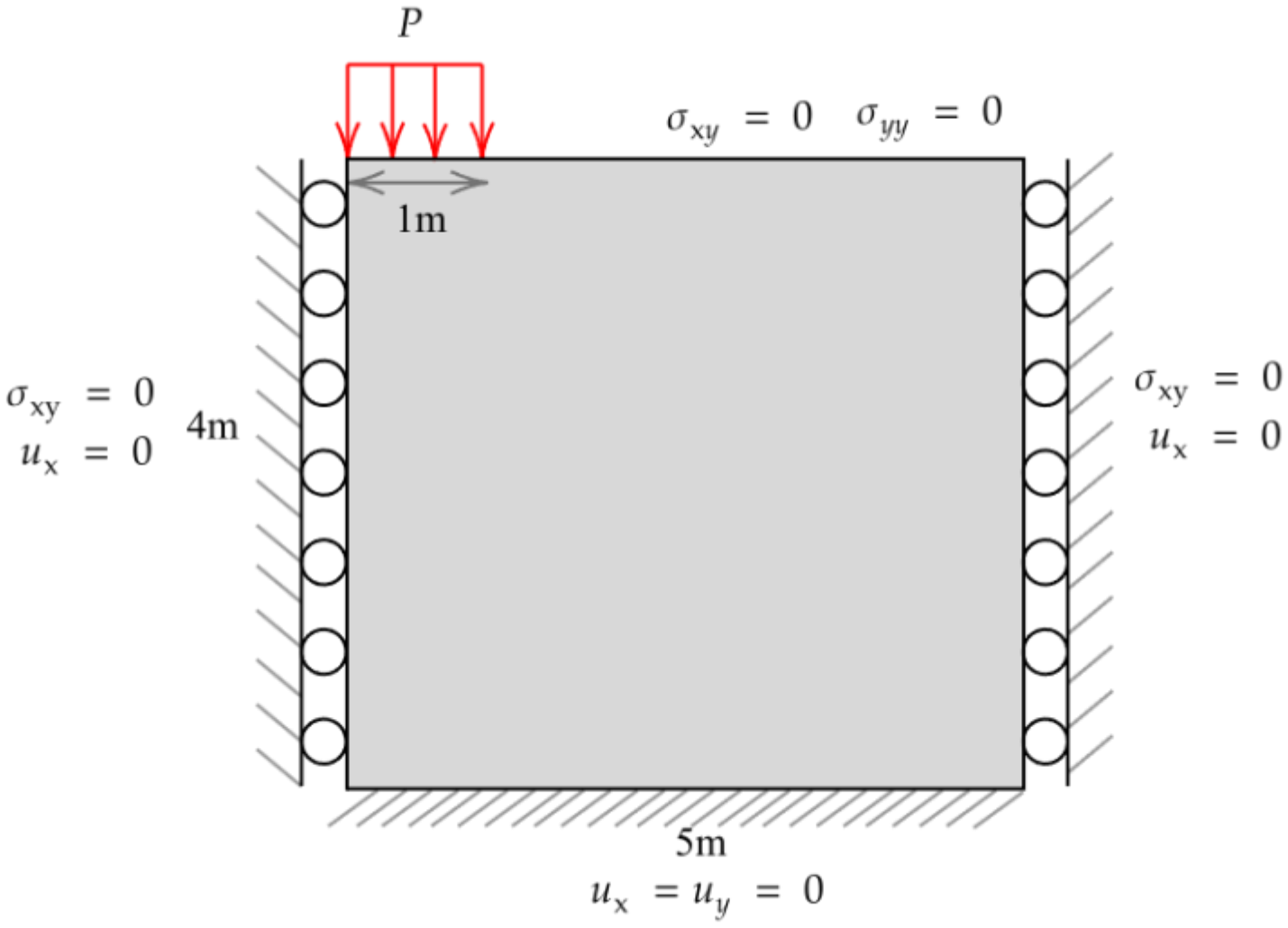

**Fig. 5.** Problem definition of the 2D domain under partial uniform loading with specified displacement constraints.

In this study, a two-dimensional plane strain rectangular domain representing a homogeneous and isotropic soil mass is considered. Body forces are neglected to simplify the numerical formulation. As illustrated in Fig. 5, a partial uniform load of magnitude P is applied over a 1 m wide strip at the top-left surface. To represent the boundary conditions, the lateral sides are modeled as a smooth roller ($u_x = 0$), while the bottom boundary is fully fixed, imposing zero displacement in both horizontal and vertical directions ($u_x = 0$, $u_y = 0$).

This configuration is selected as the benchmark problem for several reasons. First, the strip-loading arrangement constitutes a canonical bearing-capacity problem in geotechnical engineering, directly analogous to a shallow foundation resting on a soil mass, which makes the resulting stress and deformation fields physically interpretable and engineering-relevant. Second, the partial loading over a narrow strip induces a strongly non-uniform stress field with a localized plastic zone and a sharp elastic-plastic transition beneath and adjacent to the loaded edge. This is precisely the high-gradient regime in which spectral bias most severely degrades conventional PINN predictions, and it therefore provides a stringent test of the proposed Fourier feature mapping. Third, the asymmetry introduced by applying the load to only one portion of the surface produces a spatially heterogeneous response that prevents trivial symmetric solutions and demands accurate resolution of both displacement and plastic strain across the domain. Finally, the simple rectangular geometry with well-defined boundary conditions admits a reliable high-fidelity FEM reference solution, enabling an unambiguous quantitative comparison between the conventional PINN, the proposed FF-PINN, and the reference fields while isolating the effect of the constitutive nonlinearity from geometric complexity.

Table 2 lists the material properties for the three test cases. Young's modulus, Poisson's ratio, and the dilatancy angle are held constant so that differences in model performance are attributable to plastic yielding parameters alone. Case 1 serves as the baseline. Case 2 halves the cohesion and reduces the applied load, while Case 3 raises the friction angle at the baseline cohesion. The zero dilatancy angle throughout enforces a non-associative flow rule with no plastic volume change, which is representative of soils at critical state and gives the maximum departure between the yield function and the plastic potential.

**Table 2**. Material properties used in numerical examples

| Parameter | Symbol | Unit | Case 1 | Case 2 | Case 3 |
|---|---|---|---|---|---|
| Young's modulus | $E$ | kPa | 15,000 | 15,000 | 15,000 |
| Poisson's ratio | $v$ | - | 0.333 | 0.333 | 0.333 |
| Cohesion | $c$ | kPa | 20 | 10 | 20 |
| Friction angle | $\varphi$ | ° | 20 | 20 | 25 |
| Dilatancy angle | $\psi$ | ° | 0 | 0 | 0 |
| Applied load | $P$ | kPa | 120 | 50 | 110 |

Note: $E$ and $v$ are kept constant across all cases to isolate the effect of plastic yielding parameters on model performance.

### 5.2 Hyper-parameter Tuning and Sampling Strategies

#### 5.2.1 Sampling Strategies

**Table 3.** Relative $L_2$ Error of PINN and FF-PINN under Grid and Strain Adaptive Sampling (SAS)

| Variable | Relative $L_2$ Error | | | |
|---|---|---|---|---|
| | PINN (Grid) | PINN (SAS) | FF-PINN (Grid) | FF-PINN (SAS) |
| $u_x$ | 0.2000 | 0.4315 | 0.1930 | 0.1457 |
| $u_y$ | 0.0883 | 0.1673 | 0.0534 | 0.0422 |
| $\sigma_{xx}$ | 0.2582 | 0.3777 | 0.2179 | 0.1888 |
| $\sigma_{yy}$ | 0.0749 | 0.0783 | 0.0630 | 0.0518 |
| $\sigma_{zz}$ | 0.0981 | 0.1026 | 0.0952 | 0.0861 |
| $\sigma_{xy}$ | 0.2190 | 0.4060 | 0.2036 | 0.2145 |

To achieve optimal performance during training, an evaluation of sampling strategies was conducted before the architectural assessment. This study included an analysis of key hyperparameters, including the depth and width of the network, as well as the Fourier scaling parameters. A preliminary hyperparameter sweep was conducted across varying mapping sizes and scale parameters ($\sigma \in \{0.01, 0.05, 0.10\}$). To isolate the effect of network depth and width, an intermediate value of $\sigma = 0.05$ was fixed as a provisional baseline for the architecture evaluation in this section; the scale parameter is optimized separately once the optimal architecture is established. All hyperparameter tuning used 10,000 training epochs, 1,237 training data points, and 5,000 collocation points. From Table 3, the FF-PINN with Strain Adaptive Sampling (SAS) consistently achieves the lowest relative $L_2$ errors across nearly all predicted fields, outperforming the conventional PINN under both sampling strategies. A minor exception is observed for $\sigma_{xy}$, where FF-PINN (Grid) marginally outperformed FF-PINN (SAS), which may be attributed to the stochastic nature of adaptive sampling in a shear-dominated region. These results indicate that SAS consistently benefits FF-PINN more than the conventional PINN, suggesting that Fourier feature mapping provides sufficient representational capacity to effectively exploit adaptive sampling. Consequently, SAS is adopted as the fixed sampling strategy for all subsequent experiments. For this evaluation, a network of 5 hidden layers, 64 neurons per layer and the tanh activation function was used as the initial configuration, with a Fourier mapping size of 64 and $\sigma = 0.05$. All experiments were performed on a workstation equipped with an NVIDIA GeForce RTX 5090 GPU (32GB VRAM), an Intel 12$^{th}$ Generation Core processor (12 vCPUs), and 64GB of system RAM. The effect of network depth and width on model accuracy was evaluated using FF-PINN with SAS. The Fourier mapping and scale parameters were fixed at 64 and $\sigma$ = 0.05, respectively, while the number of hidden layers = {3,4,5} and neurons per layer = {32,64,128,256} were varied systematically.

Figs. 6 and 7 together isolate the source of the improvement. Fig. 6 compares the evolution of the total loss for the conventional PINN and FF-PINN under the two sampling strategies. Under grid sampling in panel (a), the two architectures follow closely similar trajectories through most of training, separating only in the later epochs. Under strain-adaptive sampling in panel (b), the separation appears early and persists, with FF-PINN reaching a lower plateau throughout. Adaptive sampling alone therefore confers limited benefit on the conventional architecture; the representational capacity supplied by the Fourier feature mapping is what allows the concentrated points to be exploited, consistent with the error trends in Table 3. Fig. 7 shows where in the domain this advantage is realized, presenting the predicted displacement and stress fields for Case 1 with the corresponding pointwise relative error maps. Each row corresponds to one field variable, with the FEM reference, the two predictions, and the two error maps arranged across the columns. Both models reproduce the overall field patterns, but the conventional PINN accumulates elevated error along the upper-left region beneath the loaded strip, most visibly in the horizontal displacement and shear stress, whereas FF-PINN leaves this region largely clear. The improvement is thus concentrated precisely in the high-gradient zone toward which adaptive sampling directs the training points, and which spectral bias otherwise prevents the standard architecture from resolving.

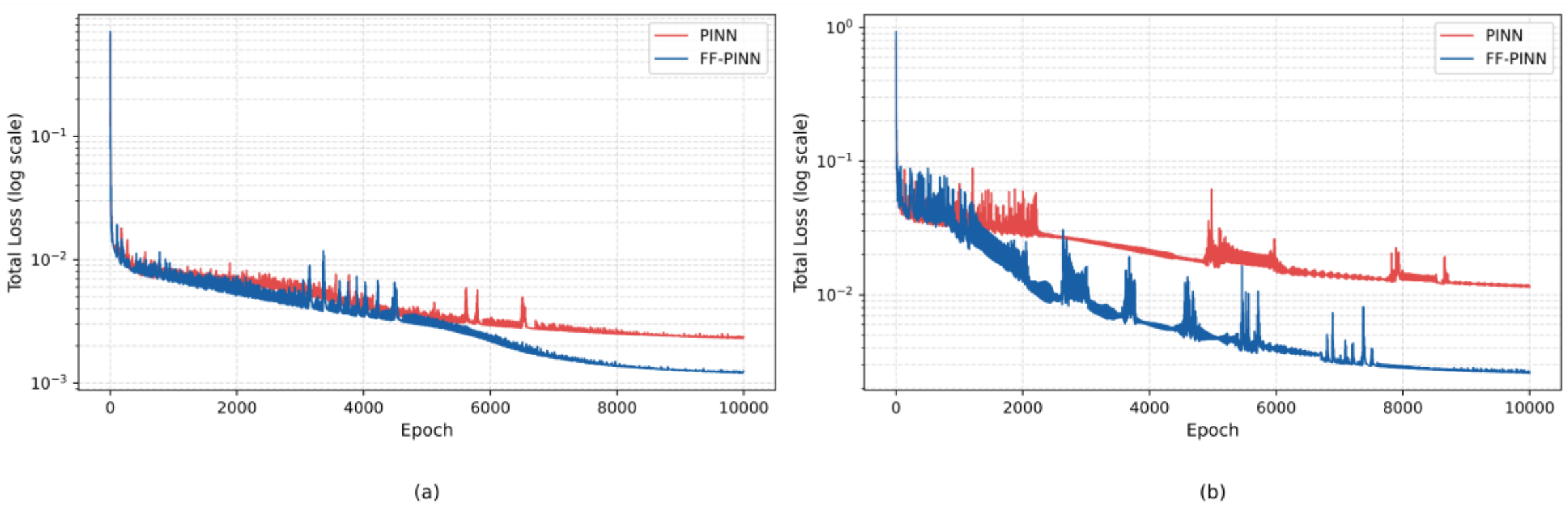

**Fig. 6.** Evolution of the total loss function during the training a) Grid Sampling, b) Strain Adaptive Sampling (SAS)

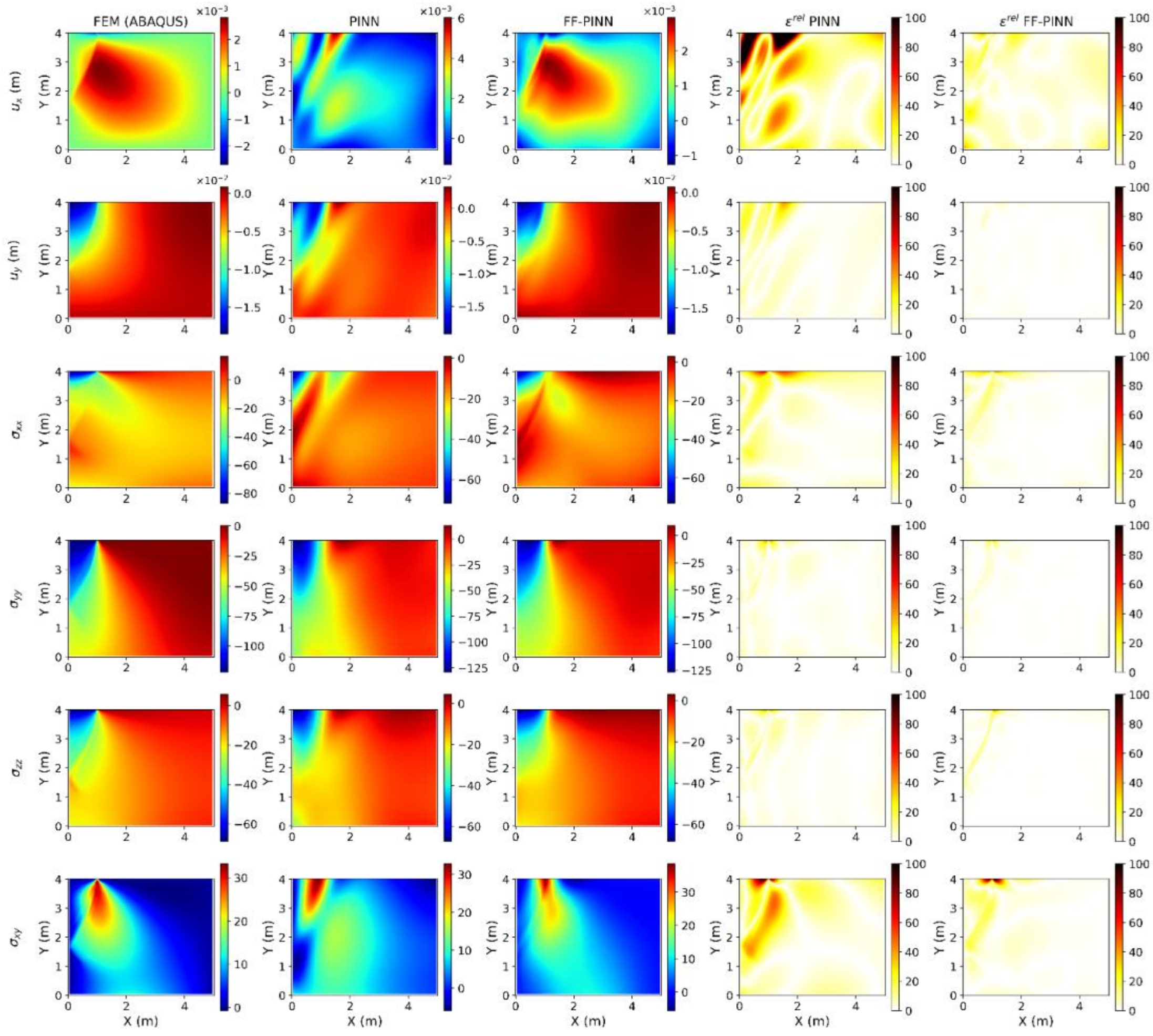

**Fig. 7.** Comparison of predicted contour plots and $\epsilon^{rel}$ for displacement and stress using strain-adaptive sampling: Conventional PINN and FF-PINN against FEM reference solutions (Case 1).

### 5.2.2 Network Architecture Size

**Table 4.** Relative $L_2$ Error for Displacements and Stresses

| [Layers, Neurons] | $u_x$ | $u_y$ | $\sigma_{xx}$ | $\sigma_{yy}$ | $\sigma_{zz}$ | $\sigma_{xy}$ |
|---|---|---|---|---|---|---|
| [3,32] | 0.8585 | 0.3580 | 0.3830 | 0.0923 | 0.1410 | 0.4545 |
| [3,64] | 0.8641 | 0.3512 | 0.4509 | 0.0998 | 0.1370 | 0.4724 |
| [3,128] | 0.7721 | 0.3624 | 0.3780 | 0.0906 | 0.1471 | 0.5030 |
| [3,256] | 0.7911 | 0.3395 | 0.3675 | 0.0836 | 0.1212 | 0.4762 |
| [4,32] | 0.4560 | 0.1539 | 0.3431 | 0.0781 | 0.1081 | 0.3302 |
| [4,64] | 0.7819 | 0.3421 | 0.3845 | 0.0895 | 0.1338 | 0.4676 |
| [4,128] | 0.7893 | 0.3037 | 0.3834 | 0.0866 | 0.1186 | 0.4642 |
| [4,256] | 0.1926 | 0.0418 | 0.2061 | 0.0592 | 0.0840 | 0.1911 |
| [5,32] | 0.2539 | 0.0946 | 0.2661 | 0.0671 | 0.0952 | 0.2757 |
| [5,64] | 0.1702 | 0.0596 | 0.2039 | 0.0583 | 0.0905 | 0.2316 |
| [5,128] | 0.2085 | 0.0554 | 0.1991 | 0.0539 | 0.0847 | 0.2026 |
| [5,256] | **0.1677** | **0.0336** | **0.1918** | **0.0423** | **0.0796** | **0.1662** |

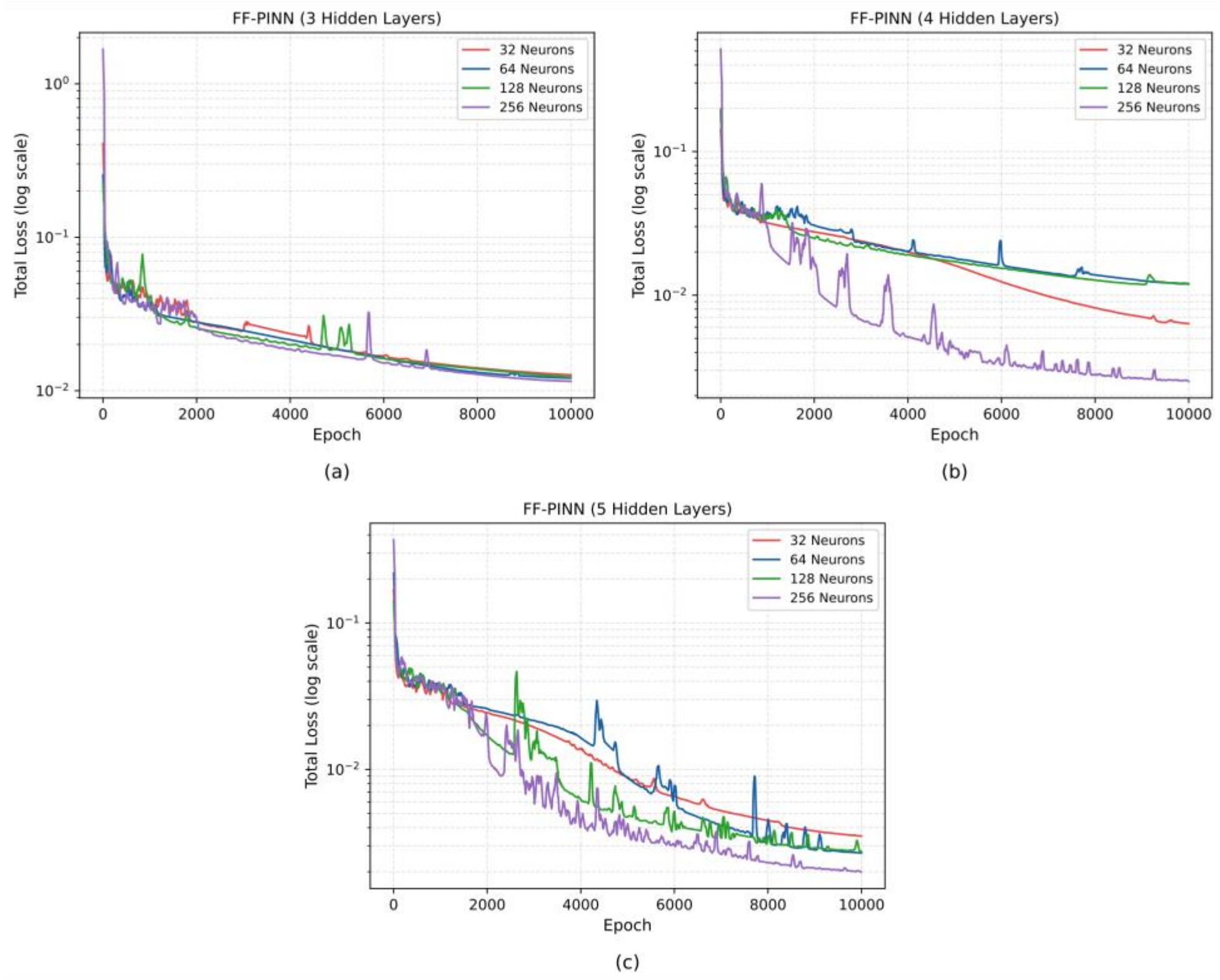

**Fig. 8.** Evolution of total loss over training epochs for FF-PINN with varying numbers of hidden layers and neurons per layer: (a) 3 hidden layers, (b) 4 hidden layers, and (c) 5 hidden layers.

Table 4 reports the relative L2 errors across a systematic sweep of network depths and widths, and Fig. 8 shows the corresponding total loss histories grouped by depth. The two are consistent in indicating that depth exerts a stronger influence than width on the capacity to capture the nonlinear elastic-plastic response. Shallow architectures exhibit persistently elevated errors regardless of neuron count, and the loss curves for three hidden layers in Fig. 8(a) collapse onto a common plateau irrespective of width, showing that additional neurons alone cannot compensate for inadequate depth. At four hidden layers in Fig. 8(b), the widest configuration separates from the remainder and converges to a markedly lower loss, while the intermediate widths remain clustered near the shallow result. This non-monotonic behavior is mirrored in Table 4, where some wider shallow configurations yield marginally higher errors than their narrower counterparts, indicating susceptibility to optimization difficulties rather than expressivity limitations. Increasing the depth to five hidden layers produces a marked and consistent reduction in error across all widths, with the loss trajectories in Fig. 8(c) descending further and faster than at either shallower depth, and the largest configuration achieving the lowest errors across all predicted fields. Across every configuration, normal stress components yield lower errors than lateral and shear stress fields, reflecting their smoother spatial gradient under the prescribed plane-strain boundary conditions. The deepest and widest configuration is therefore adopted as the fixed architecture for all subsequent experiments.

### 5.2.3 Fourier encoding layer

**Table 5.** Relative $L_2$ Error for Different Mapping Sizes and Scale Parameters

| [Mapping, $\sigma$] | $u_x$ | $u_y$ | $\sigma_{xx}$ | $\sigma_{yy}$ | $\sigma_{zz}$ | $\sigma_{xy}$ |
|---|---|---|---|---|---|---|
| [64,0.01] | 0.2774 | 0.0998 | 0.3013 | 0.0608 | 0.0921 | 0.2494 |
| [64,0.05] | 0.1677 | 0.0336 | 0.1918 | 0.0423 | 0.0796 | 0.1662 |
| [64,0.10] | 0.0881 | 0.0261 | 0.1522 | 0.0366 | 0.0695 | 0.1422 |
| [128,0.01] | 0.2931 | 0.0807 | 0.2886 | 0.0705 | 0.0953 | 0.2223 |
| [128,0.05] | 0.1493 | 0.0373 | 0.1882 | 0.0412 | 0.0750 | 0.1594 |
| [128,0.10] | 0.1381 | 0.0605 | 0.1745 | 0.0480 | 0.0806 | 0.1577 |
| [256,0.01] | 0.1862 | 0.0592 | 0.2598 | 0.0462 | 0.0871 | 0.2049 |
| [256,0.05] | 0.1381 | 0.0338 | 0.1854 | 0.0568 | 0.0825 | 0.1784 |
| [256,0.10] | 0.0771 | 0.0267 | 0.1499 | 0.0350 | 0.0652 | 0.1289 |
| [512,0.01] | 0.1573 | 0.0563 | 0.2032 | 0.0455 | 0.0813 | 0.1666 |
| [512,0.05] | 0.0750 | 0.0289 | 0.1454 | 0.0351 | 0.0680 | 0.1435 |
| [512,0.10] | **0.0578** | **0.0208** | **0.1171** | **0.0330** | **0.0680** | **0.1186** |

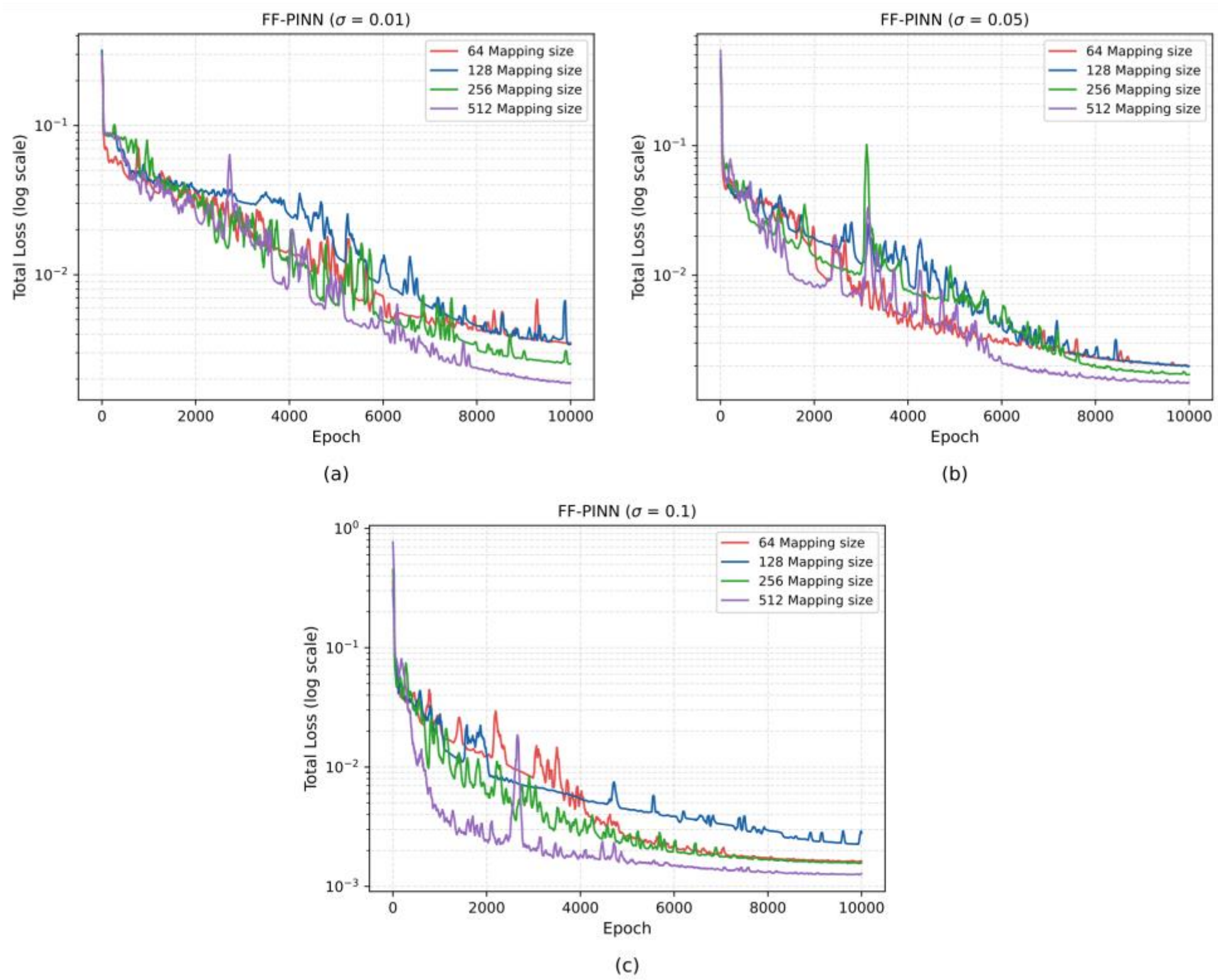

**Fig. 9.** Evolution of total loss over training epochs for FF-PINN with varying numbers of mapping size and scale parameters: (a) $\sigma = 0.01$ , (b) $\sigma = 0.05$, and (c) $\sigma = 0.10$

Table 5 summarizes the relative $L_2$ errors for varying Fourier mapping sizes = {64,128,256,512} and scale parameters $\sigma$ = {0.01, 0.05, 0.10} with the optimal architecture fixed from Section 5.2.2. The results demonstrate that the scale parameter $\sigma$ exerts a more dominant influence on model performance than the mapping size alone. Increasing $\sigma$ consistently reduces errors across all field variables and mapping sizes, reflecting its role in controlling the frequency bandwidth of the random Fourier features: a larger $\sigma$ enables the network to represent higher-frequency spatial variations inherent in the elasto-plastic fields. Although increasing the mapping size yields marginal improvements in accuracy, its effect is subordinate to that of $\sigma$, with diminishing returns observed at higher values, particularly for the smoother stress component. The best-performing configuration combines the largest mapping size with the highest scale parameter, achieving the lowest errors across nearly all fields. These results suggest a synergistic interaction between these two hyperparameters: a denser set of random frequencies drawn from a wider distribution provides more comprehensive spectral coverage of the target fields. It is, however, acknowledged that larger mapping size increases the input dimensionality to the first hidden layer, introducing additional computational cost that warrants practical considerations. Based on these findings, the optimal Fourier configuration is adopted for all subsequent model evaluations.

Fig. 9 shows the total loss histories for the Fourier mapping sweep, grouped by scale parameter, and complements the errors reported in Table 5. At the smallest scale in panel (a), all mapping sizes descend slowly and remain closely bunched, with the encoding supplying to narrow a frequency bandwidth for any mapping size to resolve the localized plastic features. At the intermediate scale in panel (b), the curves separate and reach lower plateaus, with the largest mapping size descending furthest. At the largest scale in panel (c), the descent is both faster and deeper across all mapping sizes, and the widest mapping settles nearly an order of magnitude below the narrowest. The progression across panels is larger than the spread within any single panel, confirming the conclusion drawn from Table 5 that the scale parameter dominates the mapping size: bandwidth determines what the network can represent, while mapping density refines the spectral coverage within that bandwidth.

**Table 6.** Neural Network Configuration

| Category | Hyper-parameter | PINN | FF-PINN |
|---|---|---|---|
| Network Architecture | Hidden layers | 5 | 5 |
| | Neurons per layer | 256 | 256 |
| | Activation function | Tanh | Tanh |
| Fourier Feature | Mapping size | - | 512 |
| | Scale parameter ($\sigma$) | - | 0.10 |
| Training | Optimizer | Adam | Adam |
| | Initial learning rate | 1e-3 | 1e-3 |
| | LR Scheduler | Cosine Annealing | Cosine Annealing |
| | Min. learning rate | 1e-4 | 1e-4 |
| | Batch-size | Full-batch | Full-batch |
| Sampling | Collocation strategy | SAS | SAS |

Table 6 summarizes the training configuration adopted for both models. The two share an identical backbone of five hidden layers with 256 neurons per layer and tanh activation, the configuration established in Section 5.2.2, and differ only in the Fourier feature mapping applied at the input, for which FF-PINN uses a mapping size of 512 and a scale parameter of 0.10 as identified in Section 5.2.3. Holding the backbone, optimizer, learning rate schedule, batch strategy, and sampling scheme fixed across both models ensures that any difference in performance is attributable to the input encoding alone rather than to network capacity or optimization settings. Both models were trained with the Adam optimizer (Kingma and Ba, 2014) at an initial learning rate of 1e-3, annealed to a minimum of 1e-4 under a cosine schedule (Loshchilov and Hutter, 2016), in full-batch mode with strain-adaptive collocation sampling. The cosine schedule permits rapid early descent while progressively reducing the step size, which stabilizes the later stages of training where the sharp gradients at the elastic-plastic boundary dominate the residual loss.

### 5.3 PINN and FF-PINN Results

**Table 7.** Comparison of relative $L_2$ errors between PINN and FF-PINN across different test cases.

| Field | PINN | FF-PINN | PINN | FF-PINN | PINN | FF-PINN |
|---|---|---|---|---|---|---|
| | Case 1 | | Case 2 | | Case 3 | |
| $u_x(m)$ | 0.0793 | 0.0388 | 0.0657 | 0.0257 | 0.0333 | 0.0375 |
| $u_y(m)$ | 0.0231 | 0.0143 | 0.0235 | 0.0080 | 0.0099 | 0.0067 |
| $\sigma_{xx}(kPa)$ | 0.1566 | 0.1172 | 0.0330 | 0.0285 | 0.0413 | 0.0380 |
| $\sigma_{yy}(kPa)$ | 0.0370 | 0.0280 | 0.0138 | 0.0101 | 0.0162 | 0.0124 |
| $\sigma_{zz}(kPa)$ | 0.0667 | 0.0630 | 0.0162 | 0.0167 | 0.0197 | 0.0175 |
| $\sigma_{xy}(kPa)$ | 0.1428 | 0.1060 | 0.0443 | 0.0418 | 0.0503 | 0.0464 |
| $\varepsilon_{xx}$ | 0.3523 | 0.2779 | 0.1867 | 0.1744 | 0.1714 | 0.1491 |
| $\varepsilon_{yy}$ | 0.1525 | 0.1113 | 0.0599 | 0.0545 | 0.0480 | 0.0397 |
| $\varepsilon_{xy}$ | 0.2150 | 0.1497 | 0.0818 | 0.0680 | 0.0602 | 0.0513 |
| $\varepsilon_{xx}^{p}$ | 0.1241 | 0.1254 | 0.4097 | 0.2996 | 0.3153 | 0.3501 |
| $\varepsilon_{yy}^{p}$ | 0.1314 | 0.0839 | 0.2735 | 0.1786 | 0.3291 | 0.2312 |
| $\varepsilon_{zz}^{p}$ | 0.6521 | 0.6130 | 0.4000 | 0.2774 | 0.6098 | 0.6378 |
| $\varepsilon_{xy}^{p}$ | 0.0939 | 0.0897 | 0.3471 | 0.4258 | 1.0680 | 0.8689 |

Note: All test cases were trained until convergence was achieved.

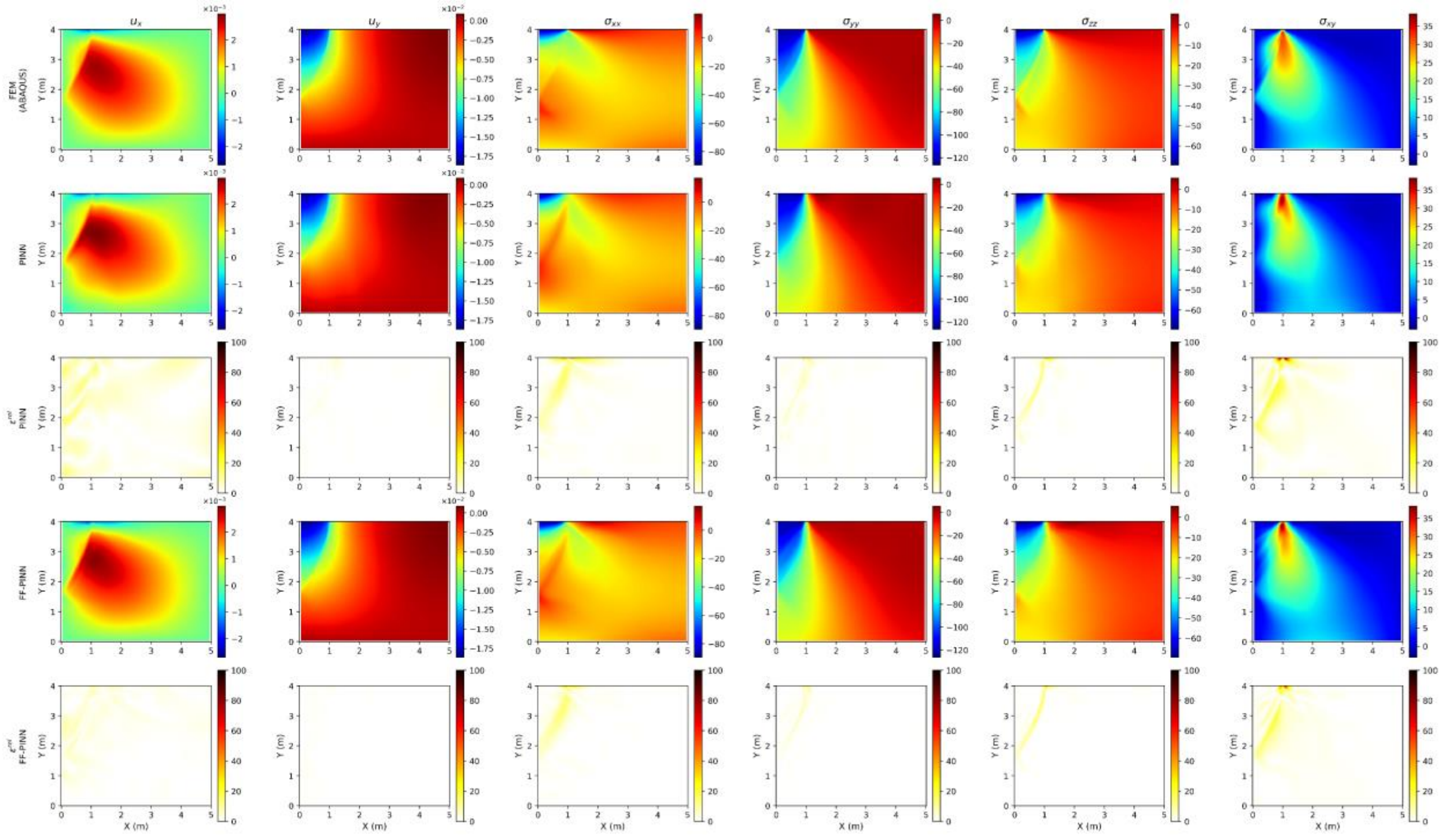

**Fig. 10.** Comparison of predicted contour plots and normalized relative errors $\epsilon^{rel}$ for displacement and stress fields using strain-adaptive sampling (SAS): FEM (ABAQUS), Conventional PINN, and FF-PINN predictions for Case 1.

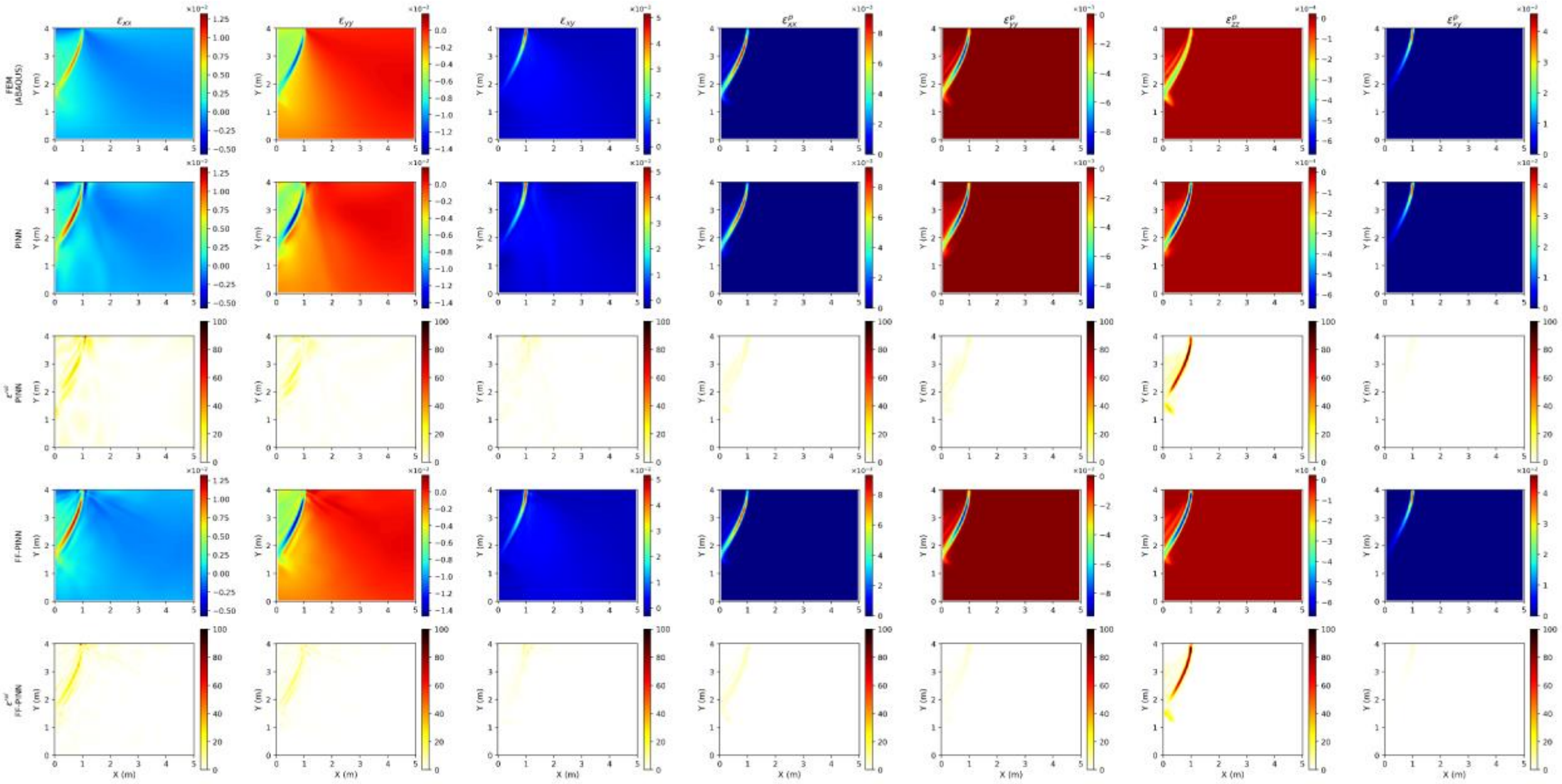

**Fig. 11.** Comparison of predicted contour plots and normalized relative errors $\epsilon^{rel}$ for total strain and plastic strain fields using strain-adaptive sampling (SAS): FEM (ABAQUS), Conventional PINN, and FF-PINN predictions for Case 1.

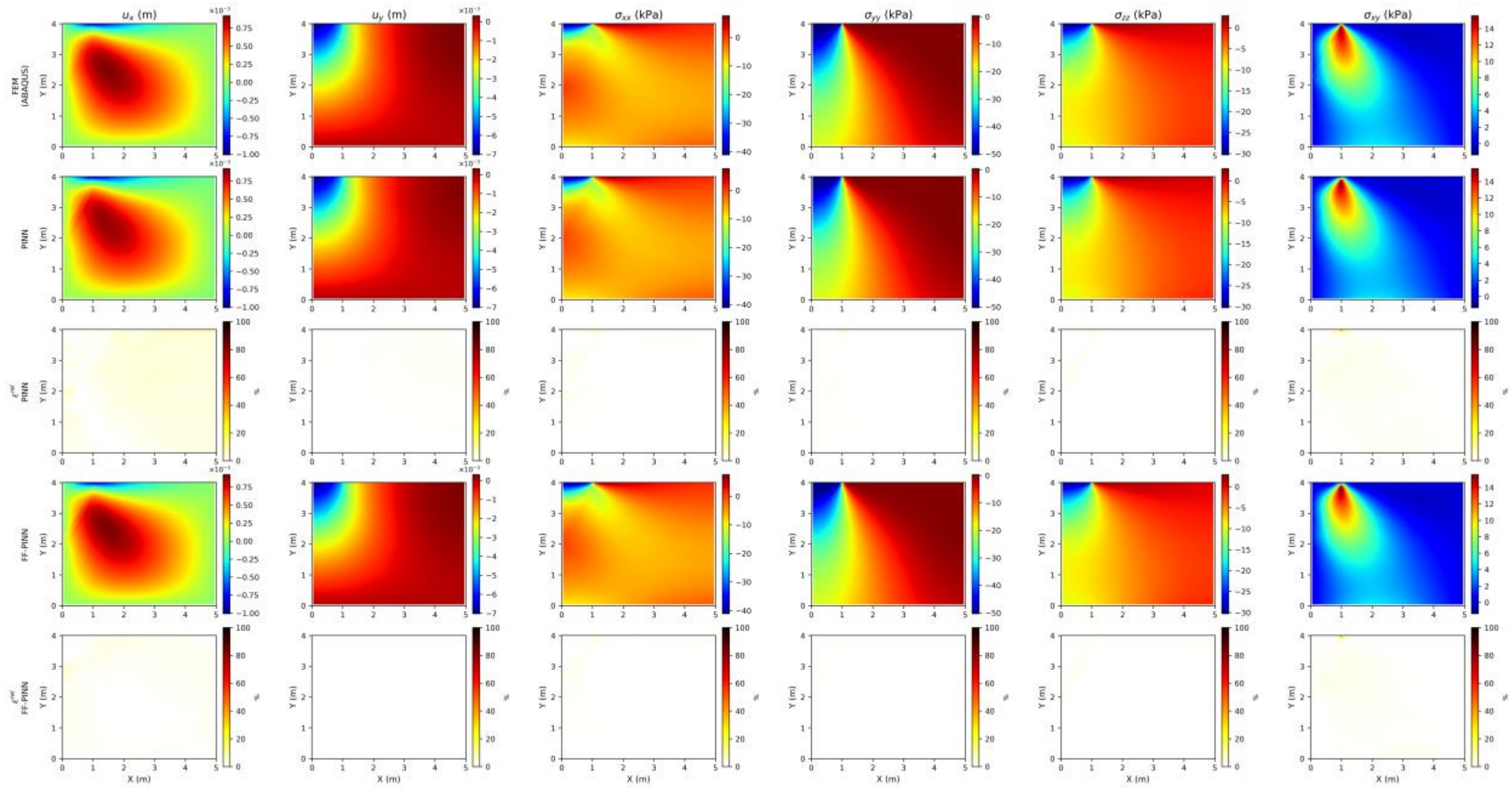

**Fig. 12.** Comparison of predicted contour plots and normalized relative errors $\epsilon^{rel}$ for displacement and stress fields using strain-adaptive sampling (SAS): FEM (ABAQUS), Conventional PINN, and FF-PINN predictions for Case 2.

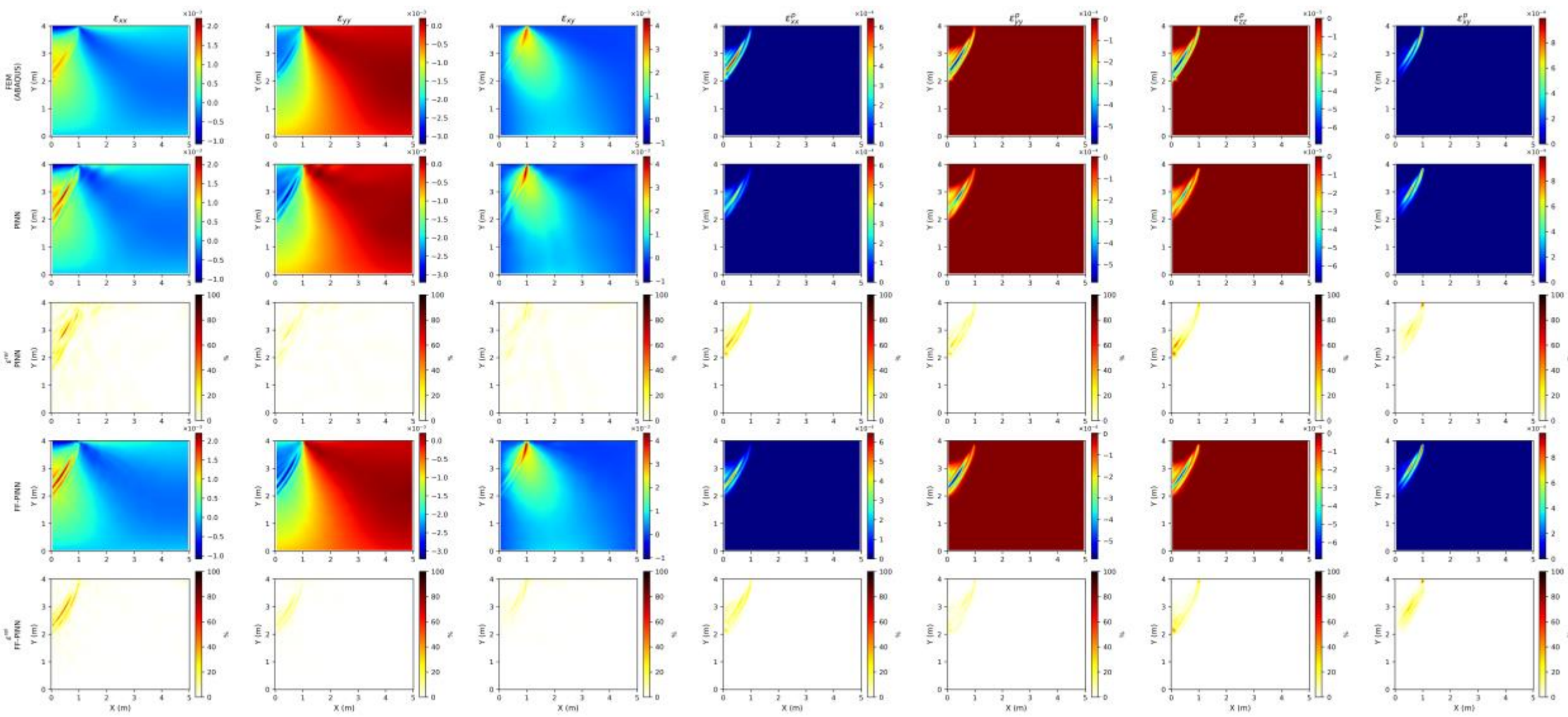

**Fig. 13.** Comparison of predicted contour plots and normalized relative errors $\epsilon^{rel}$ for total strain and plastic strain fields using strain-adaptive sampling (SAS): FEM (ABAQUS), Conventional PINN, and FF-PINN predictions for Case 2.

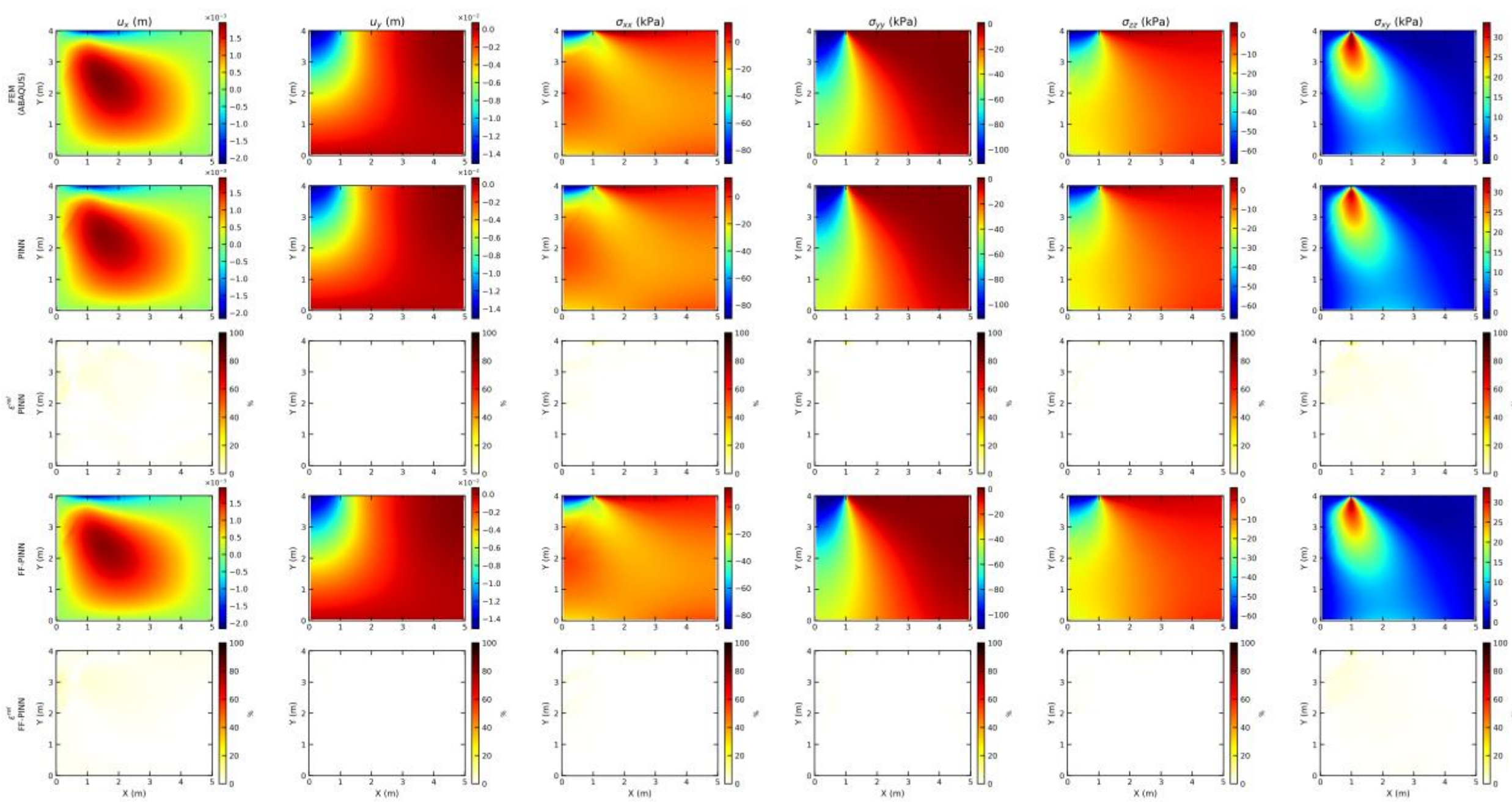

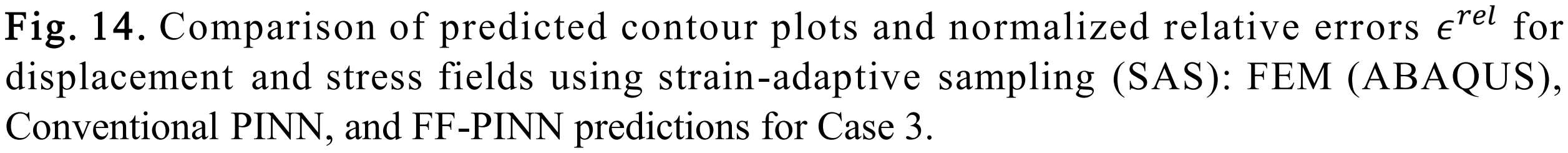

**Fig. 14.** Comparison of predicted contour plots and normalized relative errors $\epsilon^{rel}$ for displacement and stress fields using strain-adaptive sampling (SAS): FEM (ABAQUS), Conventional PINN, and FF-PINN predictions for Case 3.

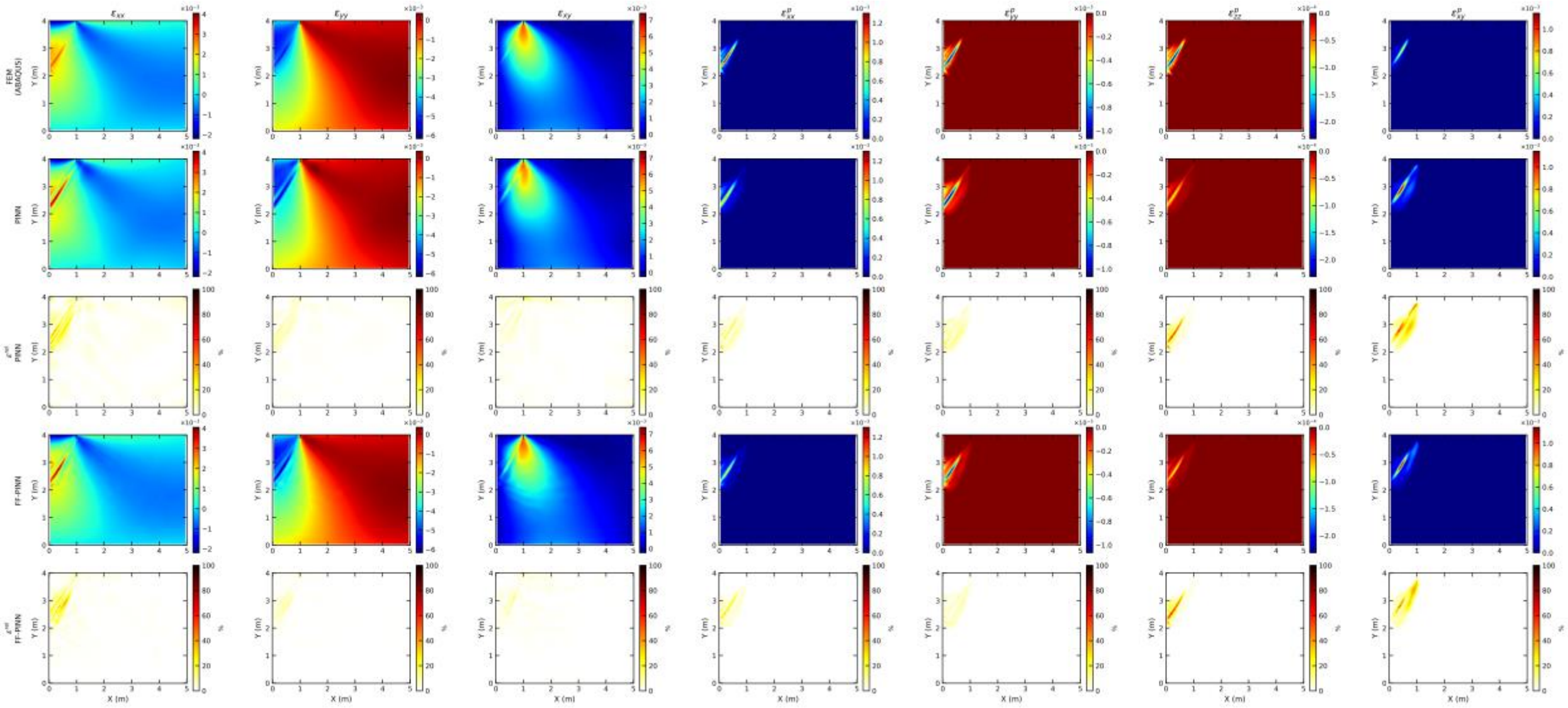

**Fig. 15.** Comparison of predicted contour plots and normalized relative errors $\epsilon^{rel}$ for total strain and plastic strain fields using strain-adaptive sampling (SAS): FEM (ABAQUS), Conventional PINN, and FF-PINN predictions for Case 3.

Figures 10 to 15 present the predicted contour plots and normalized relative errors for the displacement, stress, total strain, and plastic strain fields across all three test cases, comparing the conventional PINN, FF-PINN, and the FEM reference solutions, with the corresponding quantitative comparison summarized in Table 7. Overall, FF-PINN outperforms the conventional PINN across most field variables and test cases, with marginally higher errors confined to a small number of fields. For the displacement fields, FF-PINN reduces the error in $u_x$ by approximately 51% and 61% in Case 1 and Case 2, respectively, and in $u_y$ by approximately 38%, 66%, and 32% in Cases 1, 2, and 3, respectively. A minor exception occurs for $u_x$ in Case 3, where FF-PINN yields a slightly higher error than the conventional PINN (0.0375 versus 0.0333). Among the stress components, the largest reduction occurs for $\sigma_{yy}$ in Case 2 (approximately 27%), followed by $\sigma_{xy}$ in Case 1 (approximately 26%) and $\sigma_{xx}$ in Case 1 (approximately 25%), with $\sigma_{zz}$ in Case 2 again a notable exception where FF-PINN yields a slightly higher error than the conventional PINN. The normal stress component $\sigma_{yy}$ consistently yields the lowest relative $L_2$ errors across all cases for both models, likely reflecting its smoother and more dominant spatial distribution under the applied partial uniform vertical loading.

The improvement of FF-PINN becomes more evident in the strain and plastic strain fields. While both models produce comparable predictions in elastic dominant regions, the conventional PINN exhibits substantially higher errors in localized plastic zones where sharp spatial transitions occur. FF-PINN effectively captures these high gradient features in most cases, as reflected in the reduced relative $L_2$errors for $\varepsilon_{xx}^p$, $\varepsilon_{yy}^p$, $\varepsilon_{zz}^p$, and $\varepsilon_{xy}^p$ and in the improved agreement with the FEM solutions in Figures 11, 13, and 15. Minor exceptions occur for $\varepsilon_{xx}^p$ in Cases 1 and 3 and for $\varepsilon_{zz}^p$ in Case 3, where FF-PINN yields slightly higher errors than the conventional PINN, reflecting the inherent difficulty of resolving these highly localized plastic strain components. This overall enhancement is attributed to the Fourier feature mapping, which mitigates the spectral bias inherent in standard fully connected networks and endows the model with greater capacity to resolve the higher frequency spatial variations characteristic of localized plastic flow.

Among all predicted field variables, the plastic shear strain $\varepsilon_{xy}^p$ presents the greatest challenge for both architectures, with the conventional PINN yielding relative $L_2$ errors of 0.3471 and 1.0680 in Cases 2 and 3, respectively. FF-PINN reduces the Case 3 error to 0.8689 (approximately 19%) but yields a higher error than the conventional PINN in Case 2. Similarly, $\varepsilon_{zz}^p$ exhibits persistently elevated errors across all cases for both models, representing a shared approximation challenge associated with the out of plane plastic strain component. Under plane strain conditions, although the total out of plane strain vanishes, the plastic strain component $\varepsilon_{zz}^p$ remains nonzero to satisfy volumetric compatibility, producing a spatially complex distribution that both architectures struggle to capture accurately. This behavior is consistent with the findings of Haghighat et al. (2021), who reported elevated relative errors in the out of plane plastic strain component $\varepsilon_{zz}^p$when applying PINN to von Mises elastoplasticity, attributing the degraded accuracy to the spectral bias of standard neural network architectures in capturing highly localized, high gradient plastic deformation fields.

Figure 16 compares the predicted failure zones, defined by the boundary separating the plastic and elastic domains, obtained from FEM, the conventional PINN, and FF-PINN for the three test cases. The conventional PINN reproduces the overall extent of the plastic domain but tends to smooth its boundary, blurring the sharper lobes and localized extensions present in the FEM reference, an effect that is most apparent in Case 2 and Case 3, where the plastic zone develops a more irregular, multilobed geometry. FF-PINN reproduces these boundary features with markedly closer fidelity to FEM in all three cases, consistent with the lower plastic strain errors reported in Table 7 and discussed above. This closer geometric agreement indicates that the improvement conferred by the Fourier feature mapping is not confined to a reduction in aggregate error but extends to a more faithful recovery of the physically meaningful shape of the yielded region.

Figure 17 presents the training histories of the physics based (PDE), boundary condition (BC), and data loss components for FF-PINN and the conventional PINN across the three cases, plotted on a logarithmic scale. In all three cases, FF-PINN (solid lines) converges to lower final loss values than the conventional PINN (dashed lines) for every loss component, with the largest gap occurring in the PDE loss, which enforces the governing elastoplastic equilibrium and constitutive equations at the collocation points. The conventional PINN also displays more pronounced oscillation during training, particularly in Case 3, whereas FF-PINN follows a smoother trajectory and reaches a stable plateau within roughly half the training epochs required by the conventional PINN, consistent with the approximately 50 percent reduction in training time reported for the framework overall. These training dynamics indicate that the Fourier feature mapping improves not only the final accuracy of the predicted fields but also the stability and efficiency of the optimization process, likely because it alleviates the spectral bias that otherwise slows convergence of the high frequency loss components associated with localized plastic flow.

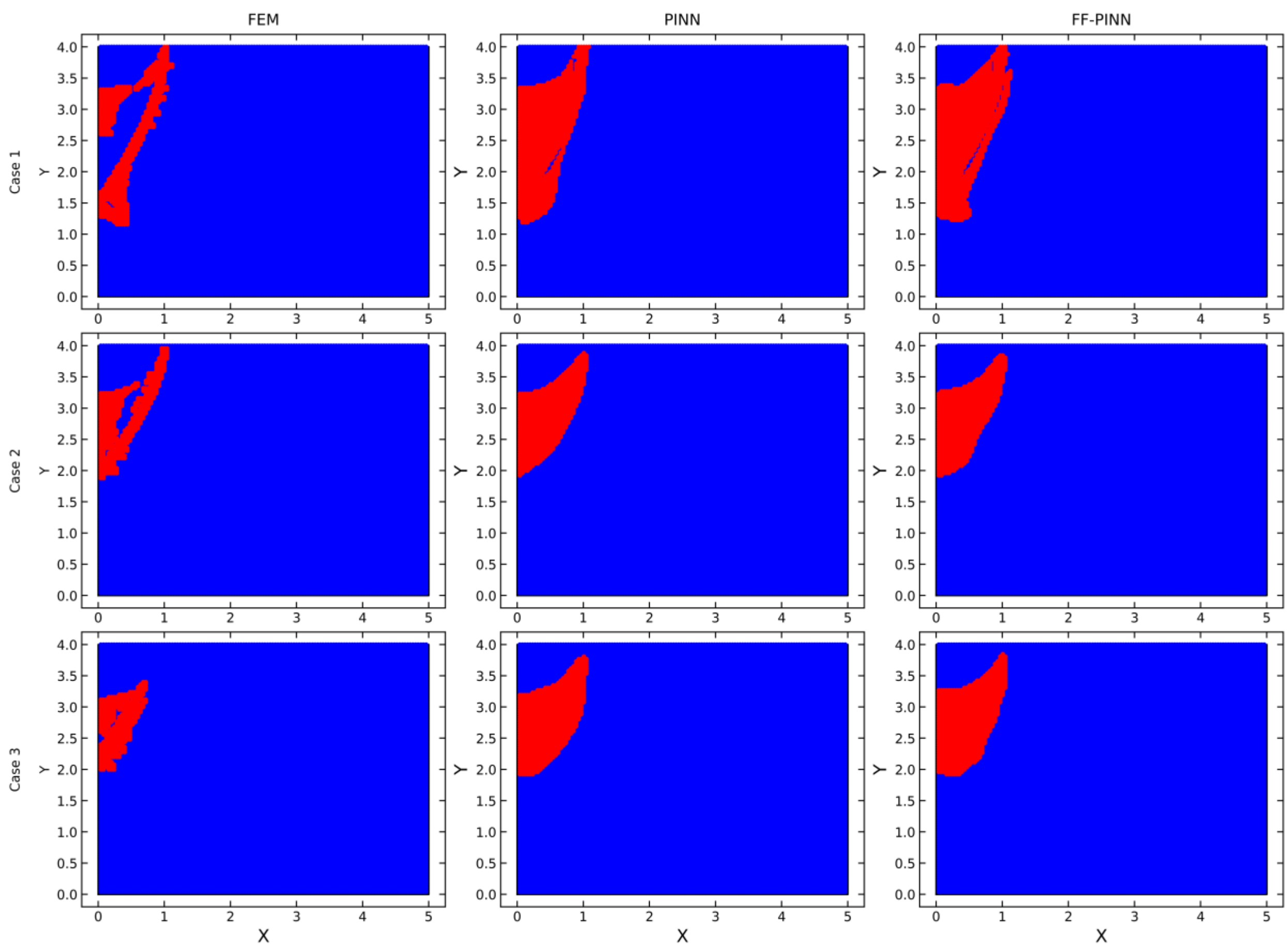

**Fig. 16.** Predicted failure zones obtained from FEM, PINN, and FF-PINN for three cases. Red and blue regions denote plastic and elastic domains, respectively.

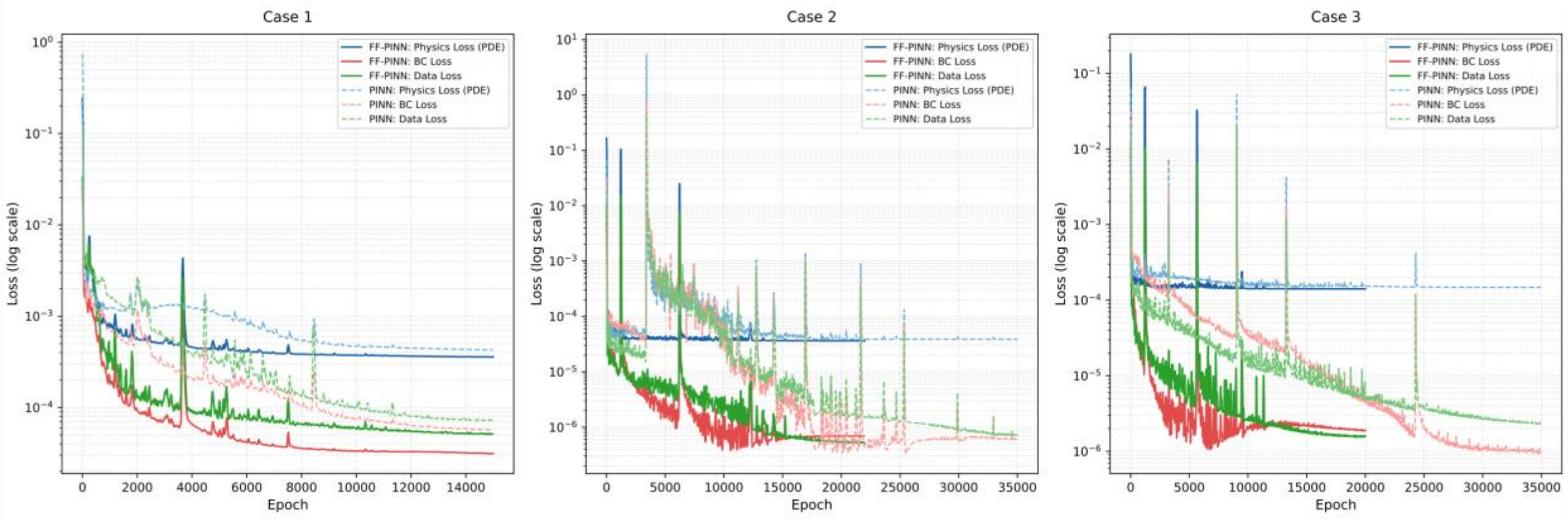

**Fig. 17.** Training history of FF-PINN vs. standard PINN for Case 1, 2, and 3. The curves represent the Physics-based (PDE), Boundary Condition (BC), and Data loss components on a logarithmic scale. Solid lines denote FF-PINN, while dashed lines represent the PINN.

### 5.4 Sensitivity analysis

In this section, the robustness and generalizability of the proposed FF-PINN framework are evaluated by examining the influence of key training parameters on model performance. The analysis focuses on four aspects: the number of training data points, the number of collocation points, the loss weighting coefficients, and the effect of noise in the training data. All experiments are performed using Case 1 as the reference configuration, with the optimal hyperparameter settings established in Table 6 unless otherwise stated.

#### 5.4.1 Influence of dataset size

Figure 18 illustrates the effect of dataset size on the %Relative $L_2$ errors across all predicted field variables, evaluated over a range from 64 to 12,500 training data points. The results demonstrate a consistent convergence pattern: % Relative $L_2$ errors decrease rapidly as the dataset grows from 64 to 1,237 data points, after which further increase yields only marginal improvement, indicating that the FF-PINN model reaches effective convergence at a moderate data volume.

For displacement fields in Fig.18(a), $u_x$ begins at approximately 10.5% error at 64 data points, decreasing rapidly before stabilizing around 4-5% for dataset sizes beyond 1,237 points with minor oscillations. The vertical displacement $u_y$ demonstrates faster convergence, dropping from approximately 4% to near 1.5-2% and remaining stable throughout, reflecting its smooth spatial distribution under the applied vertical loading. Among stress components in Fig.18(b), $\sigma_{zz}$ and $\sigma_{xx}$ exhibit the highest initial errors at approximately 23% and 19% at 64 data points, respectively, before converging to near 7% and 10% beyond 1,237 points. Meanwhile, $\sigma_{yy}$ consistently achieves the lowest errors across all dataset sizes, stabilizing near 3%, reflecting its smoother spatial gradient under the prescribed plane-strain boundary conditions. $\sigma_{xx}$ and $\sigma_{xy}$ stabilize at approximately 10%. Total strain fields in Fig. 18(c) exhibit higher absolute errors compared to displacement and stress fields. $\varepsilon_{xx}$ begins at approximately 33% and stabilizes around 26-28% for larger datasets, while $\varepsilon_{yy}$ and $\varepsilon_{xy}$ converge more favorably to approximately 12% and 15%, respectively, reflecting the greater spatial complexity of strain fields in the plastic transition zone.

The plastic strain components in Fig.18(d), present the most contrasting behavior. While $\varepsilon_{xx}^{p}$, $\varepsilon_{yy}^{p}$, and $\varepsilon_{xy}^{p}$ converge rapidly from approximately 45-51% at 64 points to near 10-15% beyond 1,237 data points, $\varepsilon_{zz}^{p}$ exhibits persistently elevated errors ranging from 60-80% across the entire dataset range with no consistent downward trend, confirming that this component poses a fundamental approximation challenge regardless of data volume, as discussed in Section 5.3.

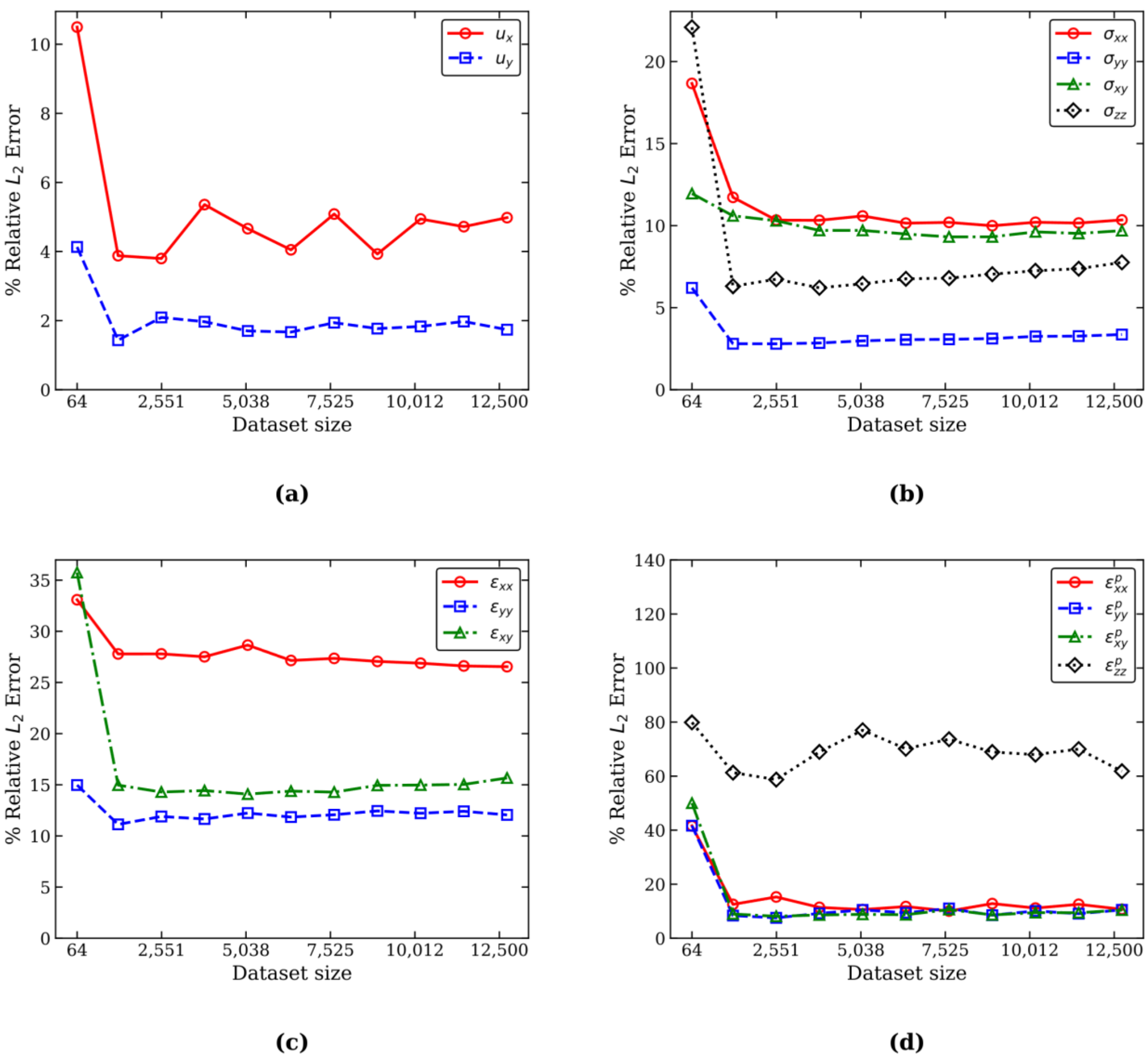

**Fig. 18.** Effect of training dataset size on the percentage relative $L_2$ errors of FF-PINN predictions for (a) displacement fields $(u_x, u_y)$; (b) stress fields $(\sigma_{xx}, \sigma_{yy}, \sigma_{zz}, \sigma_{xy})$; (c) total strain fields $(\varepsilon_{xx}, \varepsilon_{yy}, \varepsilon_{xy})$; and (d) plastic strain fields $(\varepsilon^p_{xx}, \varepsilon^p_{yy}, \varepsilon^p_{zz}, \varepsilon^p_{xy})$ under Case 1 configuration with strain-adaptive sampling.

5.4.2 Influence of Number of Collocation Points

Figure 19 illustrates the effect of the number of collocation points $N_c$ on the % Relative $L_2$ error across all predicted field variables, evaluated over the range $N_c$ = 100 to 12,000. For displacement fields in Fig 19.(a), $u_x$ exhibits the strongest sensitivity, with the error decreasing sharply from 8.6% at $N_c$ = 100 to approximately 4.0% at $N_c$ = 1,000, after which the error stabilizes with only minor fluctuations. The vertical displacement $u_y$ remains consistently low (≈1-2%) throughout, reflecting its smooth spatial distribution under the applied vertical loading. Stress components in Fig. 19(b) remain largely insensitive to $N_c$ across the full range, suggesting that physics residuals governing stress equilibrium are adequately enforced even at low collocation densities. The total strain fields in Fig. 19(c) show a gradual improvement with increasing $N_c$, particularly for $\varepsilon_{xx}$ and $\varepsilon_{xy}$, reflecting the greater spatial complexity of strain fields in plastic transition zone. For plastic strain fields Fig. 19(d), $\varepsilon_{xx}^{p}$, $\varepsilon_{yy}^{p}$, and $\varepsilon_{xy}^{p}$ remain stable near 10% across all $N_c$ values, while $\varepsilon_{zz}^{p}$ exhibits persistently elevated errors across all $N_c$ values, reflecting the fundamental approximation challenge shared by both models in capturing the out-of-plane plastic strain component, as discussed in Section 5.3. These results indicate that FF-PINN achieves stable predictions across most field variables with $N_c \geq 1{,}000$, demonstrating that the strain-adaptive sampling strategy effectively reduces the minimum collocation points requirement. The baseline of $N_c = 5{,}000$ adopted in this study therefore provides a conservative and well-justified configuration.

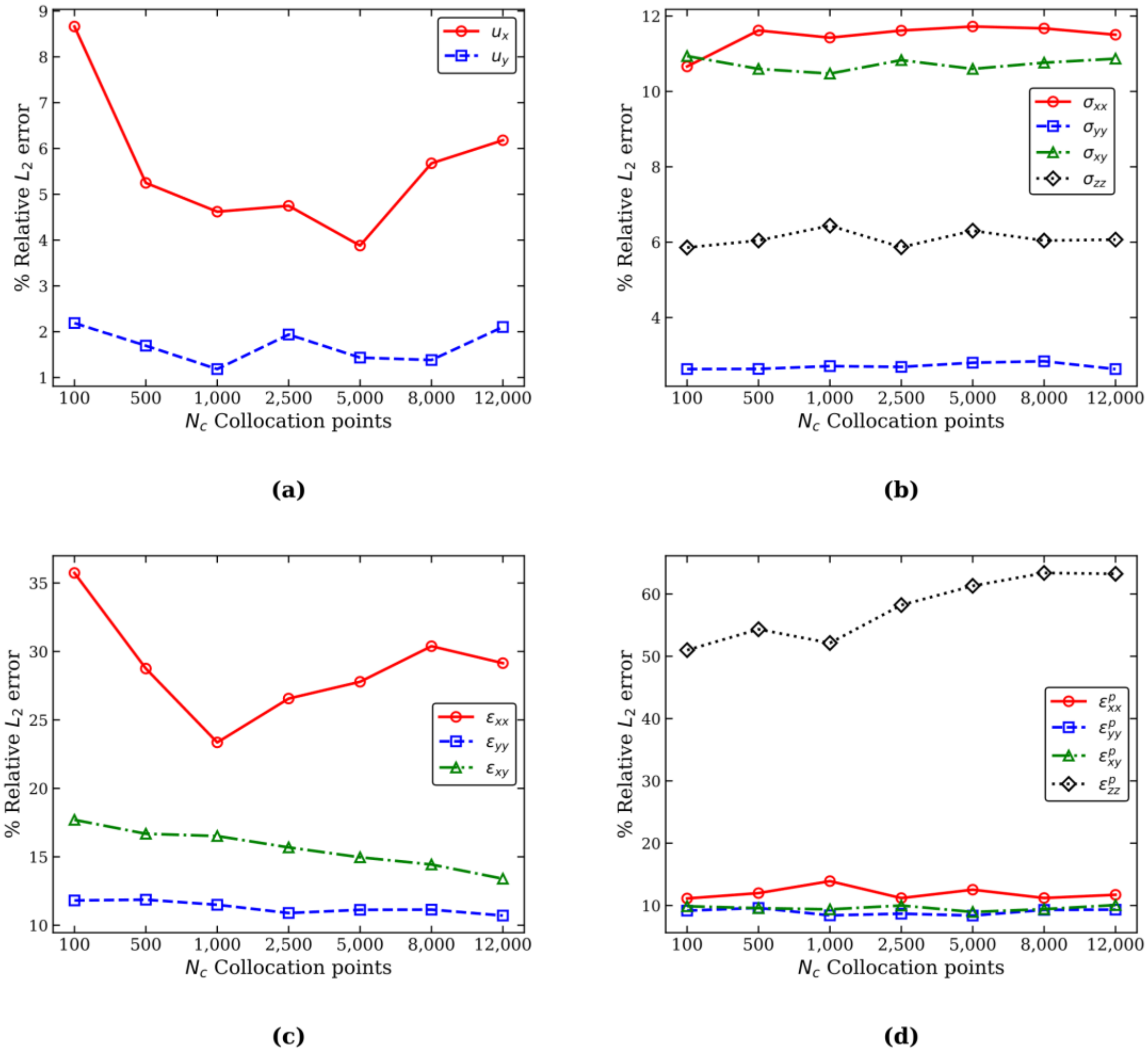

**Fig. 19.** Effect of number of collocation points $N_c$ on the % Relative $L_2$ error of FF-PINN prediction for (a) displacement fields $(u_x, u_y)$; (b) stress fields $(\sigma_{xx}, \sigma_{yy}, \sigma_{zz}, \sigma_{xy})$; (c) total strain fields $(\varepsilon_{xx}, \varepsilon_{yy}, \varepsilon_{xy})$; and (d) plastic strain fields $(\varepsilon_{xx}^{p}, \varepsilon_{yy}^{p}, \varepsilon_{zz}^{p}, \varepsilon_{xy}^{p})$ under Case 1 configuration with strain-adaptive sampling.

5.4.3 Influence of Loss Weighting Coefficients

In this section Fig. 20 presents the effect of the loss weight ratio ($w_{physics}$:$w_{data}$) on the % Relative $L_2$ error across all predicted field variables, evaluated over five configurations ranging from 1:1 to 1:30. Displacement fields in Fig. 20(a) show the clearest response to the weight ratio. The error in $u_x$ drops from 11.6% at 1:1 down to 3.9% at 1:15, then rises slightly at 1:30, suggesting that pushing $w_{data}$ too high eventually becomes counterproductive. The vertical displacement $u_y$ follows a similar but milder trend, remaining below 3% for most configurations. Stress components in Fig. 20(b) are most sensitive at the low-end; errors drop sharply between 1:1 and 1:5 but become largely insensitive beyond that point, with $\sigma_{yy}$ staying consistently near 2-3% across the upper range. Total strain fields in Fig. 20(c) respond more gradually, with $\varepsilon_{yy}$ and $\varepsilon_{xy}$ showing little variation while $\varepsilon_{xx}$ fluctuates moderately around 24-29% without a clear directional trend. Plastic strain fields in Fig.20(d) present a similar picture: $\varepsilon^p_{xx}, \varepsilon^p_{yy}, and\ \varepsilon^p_{xy}$ remain stable near 10% regardless of the weight ratio, while $\varepsilon^p_{zz}$ stays persistently high at 49-62% for reasons discussed in Section 5.3. Overall, the ratio 1:15 consistently produces the lowest or near-lowest errors across most field variables, confirming that the chosen baseline is well-justified. Configurations below 1:5 underweight the data loss and lead to noticeably degraded accuracy, while ratios beyond 1:15 offer no meaningful benefit.

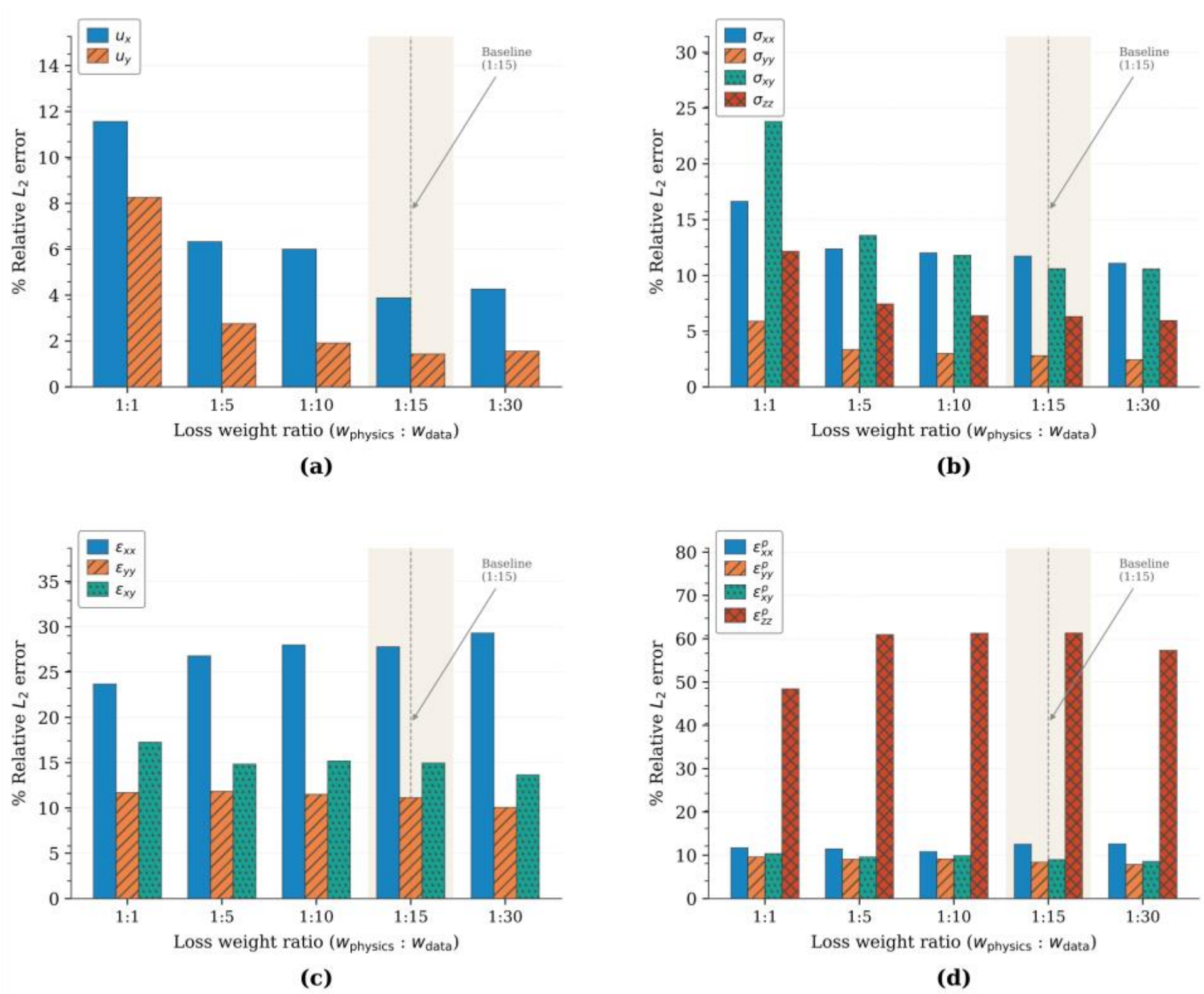

**Fig. 20.** Effect of loss weight ratio ($w_{physics}$:$w_{data}$) *on* the % Relative $L_2$ error of FF-PINN prediction for (a) displacement fields ($u_x, u_y$); (b) stress fields ($\sigma_{xx}, \sigma_{yy}, \sigma_{zz}, \sigma_{xy}$); (c) total strain

fields ($\varepsilon_{xx}, \varepsilon_{yy}, \varepsilon_{xy}$); and (d) plastic strain fields ($\varepsilon_{xx}^{p}, \varepsilon_{yy}^{p}, \varepsilon_{zz}^{p}, \varepsilon_{xy}^{p}$) under Case 1 configuration with strain-adaptive sampling.

5.4.4 Influence of Noise

To evaluate the robustness of the FF-PINN model to noisy training data, Gaussian noise was introduced into reference FEM solutions, simulating measurement uncertainties commonly encountered in practical geotechnical engineering applications such as sensor noise and field instrumentation errors. The noise is applied proportionally to the standard deviation of each field variable, defined as:

$$u_{\text{noisy}} = u_{\text{clean}} + \eta\, \sigma_{u_{\text{clean}}}\, \mathcal{N}(0,1) \quad (39)$$

where $u_{\text{clean}}$ denotes the clean reference FEM solutions, $\sigma_{u_{\text{clean}}}$ is the standard deviation of the corresponding field variable, $\mathcal{N}(0,1)$ is the standard normal distribution, and $\eta$ is the noise-level parameter, selected as 0.5%, 1.0%, 1.5% and 2.0% respectively. Fig.21 illustrates the % Relative $L_2$ error of both PINN and FF-PINN as a function of noise level η from 0.5% to 2.0%. Overall, neither model exhibits a systematic increase in mean error with increasing noise, indicating that both architectures are largely insensitive to the noise levels examined. However, a notable difference emerges in prediction stability. For total strain fields, the conventional PINN exhibits a pronounced error spike and substantially widened uncertainty band at η =1.5%, particularly for $\varepsilon_{xx}$ and $\varepsilon_{xy}$, reflecting localized training instability at this noise level. FF-PINN, by contrast, maintains a consistently flat error profile and narrow uncertainty band throughout the full noise range. A similar pattern is observed for displacement fields, where PINN shows broader variance in $u_x$ that fluctuates with noise level, while FF-PINN remains stable.

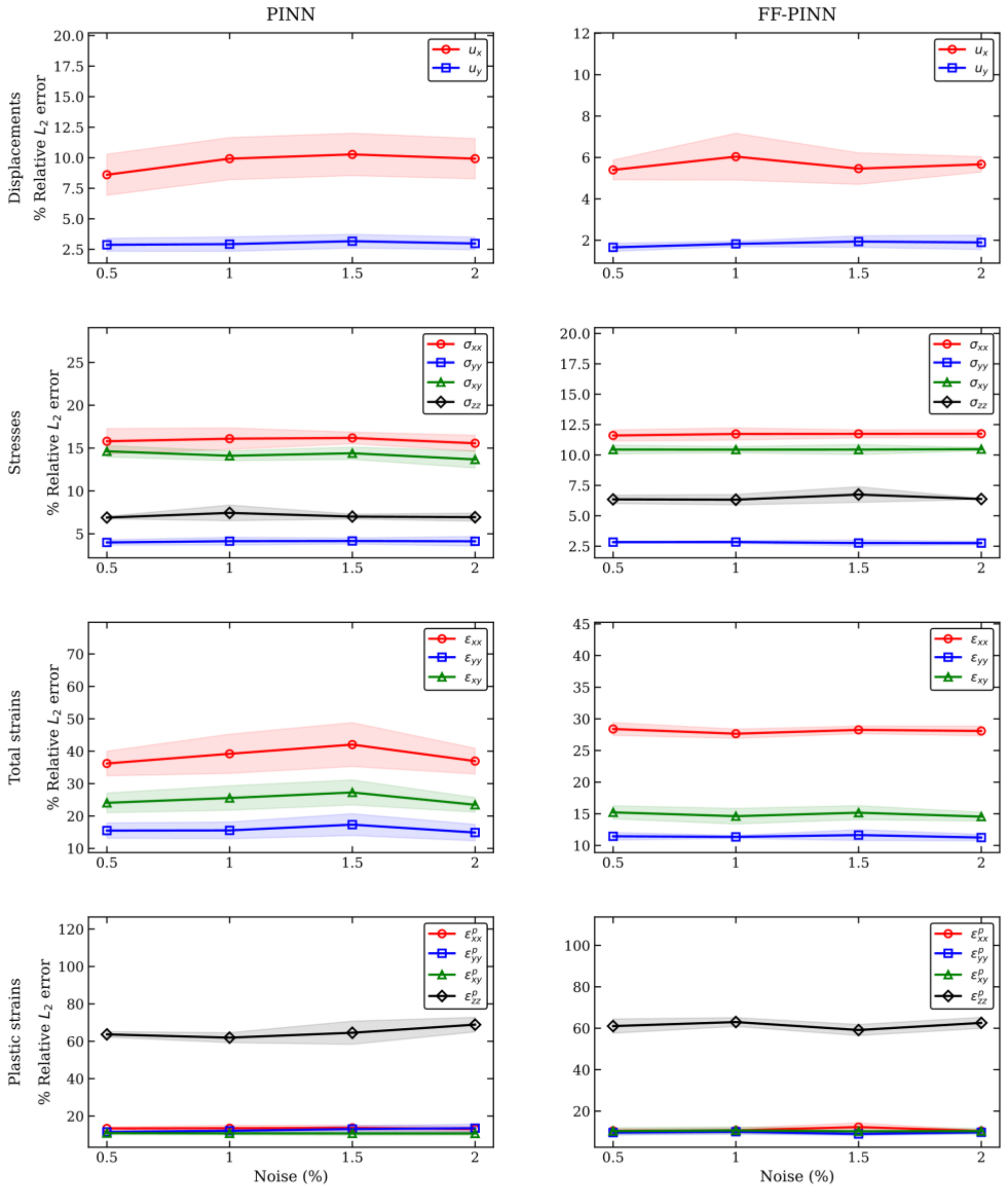

**Fig. 21.** Effect of noise level on the % Relative $L_2$ error of PINN (left column) and FF-PINN (right column) predictions for displacement fields $(u_x, u_y)$, stress fields $(\sigma_{xx}, \sigma_{yy}, \sigma_{zz}, \sigma_{xy})$, total strain fields $(\varepsilon_{xx}, \varepsilon_{yy}, \varepsilon_{xy})$, and plastic strain fields $(\varepsilon^p_{xx}, \varepsilon^p_{yy}, \varepsilon^p_{zz}, \varepsilon^p_{xy})$ under Case 1 configuration. Shaded regions represent one standard deviation across 5 independent training runs.

## 5.5. Computational cost analysis

Beyond predictive accuracy, computational cost is a decisive consideration for practical deployment. This section evaluates the training cost of the conventional PINN and FF-PINN in terms of epochs to convergence, wall-clock training time, and parameter count, using Case 2 as a representative configuration. Table 8 summarizes the comparison. FF-PINN reaches convergence at 17,500 epochs against 35,000 for the conventional PINN, halving the wall-clock training time from 1,366 s to 683 s, and does so while attaining a lower final total loss. This acceleration is obtained despite a larger trainable parameter count, 330,502 against 265,478, the increase arising from the higher input dimensionality that the Fourier mapping presents to the first hidden layer. The per-epoch cost of the two models is therefore comparable, and the saving derives from the reduced number of epochs required rather than from cheaper individual epochs. Expanding the parameter count while shortening the time to convergence indicates that the benefit lies in the optimization trajectory rather than in model size.

Fig. 22 traces the total loss histories underlying these figures and isolates the effect on convergence behavior. FF-PINN descends more rapidly from the outset and reaches a stable plateau near epoch 17,500, whereas the conventional PINN continues to oscillate through a further 17,500 epochs before settling at a comparable level. The insets make the practical consequence explicit by showing the predicted plastic shear strain contour at three points. At the matched wall-clock budget of 683 s, FF-PINN has already resolved a coherent localized plastic band, while the conventional PINN at the same elapsed time has not; only after the full 1,366 s does the conventional model recover a comparable field. The comparison at equal computational expenditure, rather than at equal epoch count, is the operative one for deployment.

This reduction in training cost does not come at the expense of accuracy. FF-PINN converges at half the epochs yet achieves substantially lower errors across most field variables in Case 2, with reductions of up to 66 percent in the vertical displacement relative to the fully trained conventional PINN (Table 7), although the out-of-plane stress and plastic shear strain components show marginally higher errors, consistent with the exceptions noted in Section 5.3. Fourier feature mapping therefore accelerates convergence while maintaining or improving accuracy, a combination that is practically significant for applications requiring repeated model evaluation, such as parametric studies or reliability analyses, where the saving compounds across runs.

**Table 8** Computational cost comparison

| Metric | PINN | FF-PINN |
|---|---|---|
| Epochs to convergence | 35,000 | 17,500 |
| Training time (s) | 1,366 | 683 |
| Trainable parameters | 265,478 | 330,502 |
| Final total loss ($\times 10^{-5}$) | 5.16 | 4.68 |

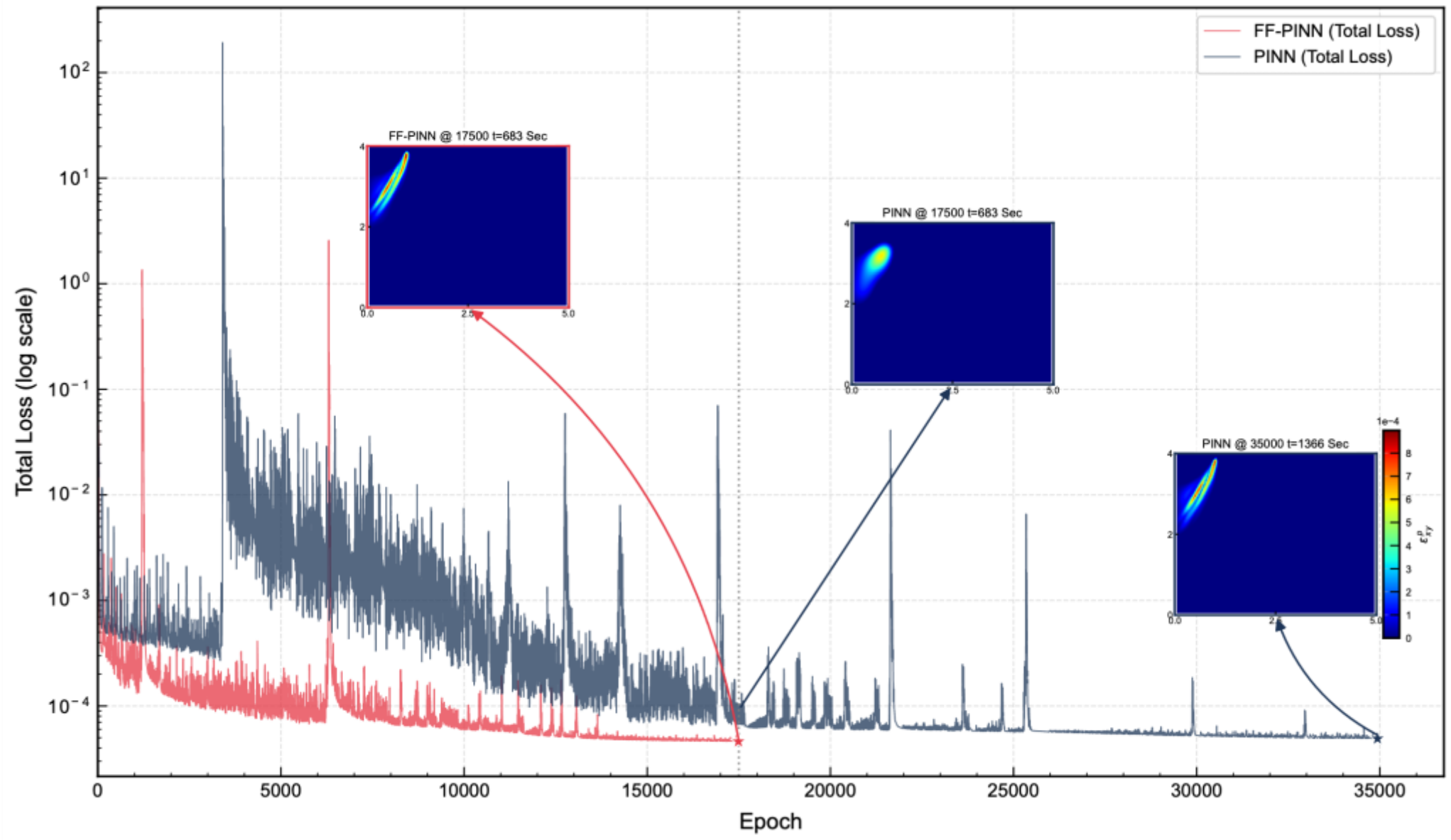

**Fig. 22.** Training loss history of PINN and FF-PINN for Case 2. Insets show the predicted $\varepsilon_{xy}^{p}$ contour at epoch 17,500 (FF-PINN convergence) and epoch 35,000 (PINN Convergence)

## 6. Discussion

The numerical results establish that embedding random Fourier feature mapping into the PINN input layer provides a consistent and substantial improvement over the conventional fully-connected architecture across most predicted field variables and test cases, with a small number of exceptions as detailed in Table 7. The most pronounced gains occur in the displacement and plastic strain fields, with relative error reductions reaching approximately 66 percent in displacement fields. This pattern directly supports the central hypothesis of the study: spectral bias is the dominant source of inaccuracy in standard PINNs applied to elasto-plastic problems, and high-frequency input encoding is an effective mechanism for resolving the sharp transitions that develop at elastic-plastic boundaries and within localized plastic zones. The faithful reproduction of the predicted plastic failure zones further confirms that the improvement is not merely a reduction in aggregate error but a genuine recovery of the physically meaningful yielding pattern.

The architectural analysis offers practical guidance for configuring the surrogate model. Network depth is shown to exert a stronger influence on accuracy than width, consistent with the view that deeper representations are required to capture the compositional nonlinearity introduced by the non-associative flow rule and the KKT activation conditions. The Fourier scale parameter emerges as a more decisive factor than the mapping size, indicating that the spectral content of the encoding, rather than its dimensionality, governs the network's ability to represent the solution. As demonstrated in Appendix A, an excessively large-scale parameter introduces spurious high-frequency content and produces spatially incoherent stress fields, which delineates a practical upper bound on this hyper-parameter and underscores the need to balance high-frequency expressivity against training stability.

The sensitivity analysis clarifies the data and sampling requirements of the framework. The relative errors converge rapidly as the dataset grows toward approximately 1,237 points and the collocation counts toward 5,000, beyond which further increases yield only marginal improvement. This convergence behavior implies that the proposed framework attains reliable accuracy at a moderate data volume, which is significant for geotechnical applications where high-fidelity reference data are costly to obtain. The loss-weight study identifies a physics-to-data ratio of 1:15 as optimal, with accuracy degrading when the physics weight is pushed too high. This finding indicates that anchoring the network to high-fidelity reference data is critical for convergence in the strongly nonlinear plastic regime, where the physics residual alone provides an insufficiently constrained optimization landscape.

The robustness assessment provides additional support for practical deployment. Neither the conventional PINN nor the FF-PINN exhibits a systematic increase in mean error as the noise level rises to 2.0 percent, demonstrating that the data-driven component does not compromise stability under the measurement uncertainties typical of field instrumentation. This insensitivity to noise is attributable to the regularizing effect of the embedded physics constraints, which suppress the propagation of stochastic perturbations into the predicted fields.

In addition to predictive accuracy, the computational cost analysis presented in Section 5.5 reveals that FF-PINN converges in approximately half the training epochs required by the conventional PINN (17,500 versus 35,000 epochs), reducing wall-clock training time from 1,366 s to 683 s despite a modest increase in trainable parameters (330,502 versus 265,478). This acceleration indicates that Fourier feature mapping not only improves solution accuracy but also guides the optimizer toward more accurate minima more efficiently, yielding simultaneous gains in both predictive performance and training economy. The result is practically significant: for problems in which repeated evaluations are required, such as parametric studies or reliability analyses, the reduced training cost compounds into substantial overall savings without any sacrifice in accuracy.

A notable limitation concerns the persistently elevated error in one plastic strain component, which remains in the range of 60 to 80 percent across the entire dataset range with no consistent downward trend. This behavior identifies a fundamental approximation challenge associated with that component rather than a deficiency that additional data can resolve, and it represents the principal avenue for further methodological refinement. Beyond this, the present study is restricted to a two-dimensional plane-strain domain under monotonic loading with constant elastic parameters. Extension to three-dimensional geometries, more complex constitutive models, cyclic and path-dependent loading, and inverse identification of material properties therefore constitutes the natural direction for future work, alongside the integration of adaptive loss-weighting strategies and transfer learning to enhance generalizability across varying material conditions.

## 7. Conclusion

This study presented a Fourier Feature Physics-Informed Neural Network (FF-PINN) framework for solving elasto-plastic boundary value problems governed by the non-associative Mohr-Coulomb constitutive model. The framework embeds random Fourier feature mapping into the network input layer, combined with a multi-objective loss enforcing equilibrium, constitutive relations, and Karush-Kuhn-Tucker conditions against high-fidelity FEM reference data, and a strain-adaptive sampling strategy. The modification is confined to the input representation, so the resulting gains are attributable to the spectral properties of the encoding rather than to network capacity or optimization settings.

Numerical benchmarking across three test cases demonstrated that FF-PINN outperforms the conventional PINN across most predicted field variables, with a small number of exceptions detailed in Table 7, achieving error reductions of up to approximately 66 percent. The improvement is concentrated in the high-gradient regions beneath the loaded strip rather than distributed uniformly and extends beyond aggregate error to the geometry of the predicted failure zone, which FF-PINN reproduces with markedly closer fidelity to FEM than the conventional architecture, whose smoothing of the plastic boundary is the signature of spectral bias.

The architectural analysis revealed that network depth governs the representation of elasto-plastic nonlinearity more strongly than width, and that the Fourier scale parameter exerts a more decisive influence on accuracy than the mapping size, indicating that the spectral bandwidth of the encoding rather than its dimensionality is the controlling factor. An excessively large-scale parameter introduces spurious high-frequency content and degrades stability, which establishes a practical upper bound on this hyper-parameter.

The sensitivity analysis confirmed that reliable predictions are attained with approximately 1,237 training data points and 5,000 collocation points, beyond which improvements are marginal, and that the framework remains stable for noise levels up to 2.0 percent, with FF-PINN maintaining a narrower uncertainty band than the conventional PINN. The loss-weighting study identified a physics-to-data ratio of 1:15 as optimal, confirming that anchoring the network to reference data is critical for convergence in the strongly nonlinear plastic regime, where the physics residual alone provides an insufficiently constrained optimization landscape. The computational cost analysis further showed that FF-PINN converges in half the training epochs of the conventional PINN, halving wall-clock training time despite a modest increase in trainable parameters and therefore delivers higher accuracy at lower cost rather than trading one against the other.

A persistently elevated error in the out-of-plane plastic strain component was identified as a fundamental approximation challenge that additional data cannot resolve and represents the principal target for further methodological refinement. Beyond this, the present study is restricted to a two-dimensional plane-strain domain under monotonic loading with constant elastic parameters. Future work will extend the framework to three-dimensional geometries, more complex constitutive models, and cyclic loading, and will explore inverse identification of material properties, adaptive loss-weighting strategies, and transfer learning to enhance generalizability across varying material conditions.

## Acknowledgments

This research was supported by the "Petchra Pra Jom Klao Master's Degree from King Mongkut's University of Technology Thonburi"

## Data availability

Data will be made available on request.

**Appendix A.** Effect of Fourier Scale Parameter on Predicted $\sigma_{xy}$ Field

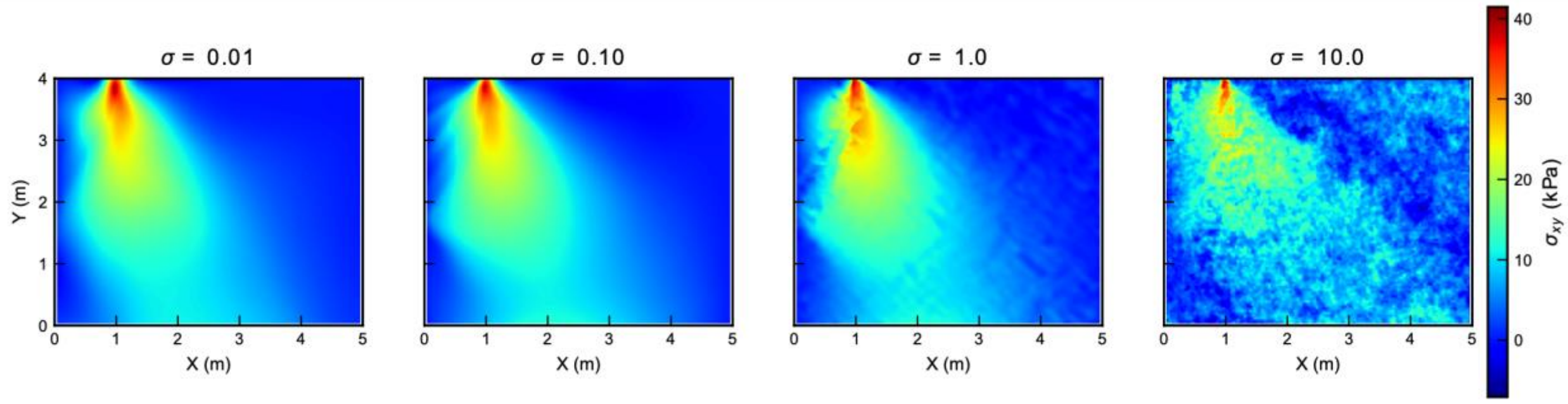

**Fig A.1.** Predicted shear stress fields from FF-PINN with varying Fourier scale parameters ( $\sigma$ = 0.01, 0.10, 1.0, and 10.0). Models with $\sigma < 1.0$ yield consistent and physically meaningful results, whereas $\sigma$ = 10.0 produces a spatially incoherent field, indicating training instability from excessive high-frequency feature mapping.

To illustrate the effect of the Fourier scale parameter on prediction quality, Fig. A.1 presents the predicted shear stress fields $\sigma_{xy}$ under each configuration. Models with $\sigma \leq 0.10$ produce spatially coherent and physically consistent fields, whereas $\sigma$ = 1.0 begins to exhibit minor oscillatory artifacts and σ = 10.0 produces a spatially incoherent field indicative of training instability from excessive high-frequency feature encoding. These results confirm that σ = 0.10 represents a practical upper bound that balances high-frequency expressivity with training stability.